 \documentclass[final,3p,times]{elsarticle}
\usepackage{amssymb}
\usepackage{color}

\usepackage{amsmath}
\usepackage{amssymb}

\usepackage{algorithm}% http://ctan.org/pkg/algorithms
\usepackage{algpseudocode}% http://ctan.org/pkg/algorithmicx

\usepackage[utf8]{inputenc}

\usepackage{afterpage}

\DeclareMathOperator*{\argmin}{arg\,min}

\begin{document}

	\newcommand{\vphi}{{\boldsymbol{\varphi}}}

	\newcommand{\W}{{\boldsymbol{W}}}

	\newcommand{\Whi}{{\W_{\hat{i}}}}
	\newcommand{\Whj}{{\W_{\hat{j}}}}
	\newcommand{\Whr}{{\W_{\hat{r}}}}
	\newcommand{\Wci}{{\W_{\check{i}}}}
	\newcommand{\Wcj}{{\W_{\check{j}}}}

	\newcommand{\vw}{{\boldsymbol{w}}}
	\newcommand{\K}{{\boldsymbol{K}}}

	\newcommand{\Q}{{\boldsymbol{Q}}}
	\newcommand{\R}{{\boldsymbol{R}}}
	\newcommand{\Rhr}{{\R_{\hat{r}}}}
	\newcommand{\E}{{\boldsymbol{E}}}

	\newcommand{\vtau}{{\boldsymbol{\tau}}}
	\newcommand{\vvarepsilon}{{\boldsymbol{\varepsilon}}}
	\newcommand{\vzero}{{\boldsymbol{0}}}
	\newcommand{\hvtau}{{\hat{\vtau}}}
	\newcommand{\M}{{\boldsymbol{M}}}
	\newcommand{\vdelta}{{\boldsymbol{\delta}}}
	\newcommand{\vchi}{{\boldsymbol{\chi}}}
	\newcommand{\vbeta}{{\boldsymbol{\beta}}}
	\newcommand{\vomega}{{\boldsymbol{\omega}}}

	\newcommand{\vq}{\mathbf{q}}
	\newcommand{\vdq}{\dot{\mathbf{q}}}
	\newcommand{\vddq}{\ddot{\mathbf{q}}}
	
	\newcommand{\vz}{\mathbf{z}}
	\newcommand{\vdz}{\dot{\mathbf{z}}}
	\newcommand{\vddz}{\ddot{\mathbf{z}}}

	\newcommand{\abs}[1]{\lvert#1\rvert}
	\newcommand{\norm}[1]{\lVert#1\rVert}
	\newcommand{\comillas}[1]{``#1''}
	\newcommand{\vect}[1]{\left\{\begin{matrix}#1\end{matrix}\right\}}
	\newcommand{\ie}{\emph{i.e.{ }}}
	
	\newcommand{\red}[1]{{\color{red}#1}}
	\newcommand{\blue}[1]{{\color{red}#1}}

    \newcommand{\sigmaunmedio}{\boldsymbol\Sigma^{1/2}}
    \newcommand{\sigmamenosunmedio}{\boldsymbol\Sigma^{-1/2}}

\begin{frontmatter}



\title{Simplification of multibody models by parameter reduction}


\author[imem]{Javier Ros\corref{cor1}\fnref{isc}}
\ead{jros@unavarra.es}

\author[imem]{Xabier Iriarte\fnref{isc}}
\ead{xabier.iriarte@unavarra.es}

\author[imem]{Aitor Plaza\fnref{isc}}
\ead{aitor.plaza@unavarra.es}

\author[upv]{Vicente Mata}
\ead{vmata@mcm.upv.es}

\cortext[cor1]{Corresponding author}

\address[imem]{Department of Mechanical Engineering, Public University of Navarre, Pamplona, Spain}
\address[isc]{Institute of Smart Cities - ISC, Pamplona, Spain}
\address[upv]{Centro de Investigaci\'on en Ingeniería Mecánica, Universidad Polit\'ecnica de Valencia, Valencia, Spain}

\begin{abstract}
Model selection methods are used in different scientific contexts to represent a characteristic data set in terms of a reduced number of parameters. Apparently, these methods have not  found their way into the literature on multibody systems dynamics. Multibody models can be considered  parametric models in terms of their dynamic parameters, and model selection techniques can then be used to express these models in terms of a reduced number of parameters. These parameter-reduced models are expected to have a smaller computational complexity than the original one and still preserve the desired level of accuracy. They are also known to be good candidates for parameter estimation purposes.

In this work, simulations of the actual model are used to define a data set that is representative of the system's standard working conditions. A parameter-reduced model is chosen and its parameter values estimated so that they minimize the prediction error on these data. To that end, model selection heuristics and normalized error measures are proposed.

Using this methodology, two multibody systems with very different characteristic mobility are analyzed. Highly considerable reductions in the number of parameters and computational cost are obtained without compromising the accuracy of the reduced model too much. As an additional result, a generalization of the base parameter concept to the context of parameter-reduced models is proposed.

\end{abstract}

\begin{keyword}
Multibody dynamics \sep Model reduction \sep Model selection \sep Base parameters \sep Parameter estimation \sep Real-time



\end{keyword}

\end{frontmatter}



%--------------------------------------------------
\section{Introduction}

The requirement of fast simulations to speed up design or to enable real-time applications has been one of the main concerns
of researchers in Multibody Systems Dynamics  (MSD)  over recent decades. This, in turn,  has
made  model reduction  a fundamental topic in the literature on MSD. 
Here, model reduction most frequently means \textit{model order reduction} (MOR), that is, a reduction
in the number of degrees of freedom (\textit{d.o.f.}) of the model without compromising its accuracy too much. A common situation
is that of \textit{linear model order reduction} of the elastic \textit{d.o.f.} of bodies. The most popular techniques include:
Component Mode Synthesis, Moment Matching Krylov subspaces, SVD-based and
Proper Orthogonal Decomposition-based techniques, etc. For an overview of these techniques see \cite{Nowakowski2012}. MOR
techniques can also be applied to full multibody systems or subsystems
using, for example, the Global Modal Parametrization \cite{Bruls2007}.

Real-time performance is a requirement in many multibody applications, such as feed forward and predictive control \cite{Cazalilla2014,Powell2015}, hybrid (analytical and data-based)
model-based control \cite{Reinhart2017}, model-based nonlinear filtering \cite{Sanjurjo2017}, 
model-based condition monitoring \cite{Charles2009,Weston2015}, etc. 
Frequently, models do not require body-level flexibility to be accounted
for and, therefore,  MOR is not suitable. However, due to the strict
demands of real-time performance,  researchers keep striving to minimize the computational
demand required by the models. In this context, formulations that solve the exact dynamic equations using a minimum number of operations as recursive
$O(n^3)$ and  $O(n)$ formulations play a predominant role \cite{Featherstone2008}.

Symbolic multibody methods can be used
to implement recursive formalisms. This reduces the operation count even more
due to the exploitation of symmetries \cite{Samin2003}. In addition to this,
the development of special inertial parametrizations such as the \textit{barycentric inertial parameters} \cite{Wittenburg2008,Samin2003}
and  \textit{minimum} or \textit{base inertial parameters} \cite{Khalil1987} should be mentioned, as these are used to further lower the operation count.
Some heuristics have also been proposed in order to determine more efficient kinematic parametrizations \cite{Leger2007}.

Other methods further lower the operation count at the expense of accuracy.
They remove or approximate some terms of the model functions, or change their computation frequency. These procedures generally rely on symbolic methods.
In \cite{Cvetkovic1982} the authors propose to remove the terms related to Coriolis and centrifugal forces and to
take out the kinematics of the terminal part from the dynamic computations; in \cite{Djurovic1991}
different terms of the dynamic model are updated using different frequencies; in \cite{Weber2003} a
statistical analysis of the values of different elements of the dynamic model functions is carried out in order
to replace some of them with suitable constants; and in \cite{Kim2007} the authors use a spline
approximation to determine the dependent coordinates as a function of the independent ones for
some closed-loop subsystems (car suspensions). Obviously, some of these strategies can be used simultaneously.

Due to the linearity of the equations of motion with respect to the inertial parameters \cite{Khosla1985}, the
typical inertia parameter estimation problem \cite{Gautier1988} takes the form of a standard linear regression problem
\begin{equation}
\W\vphi=\vchi,
\end{equation}
where $\vphi$ is the vector of inertial variables, $\vchi$ is the vector of external forces at different measurement points, and $\W$
is the observation matrix that is a function of the \textit{extended state} at the different measurement points. Standard linear model selection techniques
can be regarded as methods that ``find the subset of parameters ($\vphi$) that \textit{best} fit a given data vector ($\vchi$) as a
linear combination of some regressors (columns of $\W$)''. This selection is a combination of two elements: 1) an
error criterion to measure the accuracy of the fit \cite{Soderstrom2002}, and 2) a heuristic to select the most promising parameter
combinations \cite{Draper1966,Miller1990,Nelles2001,Furnival1974,Newton1967a}. Validation techniques which measure the error criterion for
the selected model on a validation data set are frequently used. 
%When not many data are available, some of the classic criteria
%for the best model selection are C$_p$,  FPE, AIC, BIC, and MDL \cite{Ljung1998,Myers1990,Seber2003}.
% * <jros@unavarra.es> 2016-03-30T13:12:57.857Z:
%
% Yo quitaría estaú ltima frase, no tiene mucho sentido en el presente contexto, y menos en la introducción. Podría tener sentido en un articulo similar orientado a la identificación. ¿Que opinais lo quitamos de aquí?
%
% ^.

%Model reduction techniques have not apparently being introduced in the context of multibody dynamics.
In this paper, we propose the use of  model selection techniques to obtain approximate parameter-reduced dynamic models
of multibody systems. The regression  problem is fed with data
obtained from simulations based on the exact model and model selection techniques are used  to choose
the \textit{simplest} set of parameters that fit these data for a given error tolerance.
%,
%for instance one that minimizes the expectation of the prediction error (FPE criterion). 
% * <jros@unavarra.es> 2016-03-30T13:10:53.250Z:
%
% ¿Por que es FPE, de donde vienen la F?
%
% ^.

%high/low mobility of a body in a multibody system is defined as the possibility of the body to exhibit \blue{large/small} excursions
%in its relative motion space with respect to other bodies and/or with respect to the inertial frame.

To that end,  we propose  non-dimensional normalized error measures to determine the performance of direct and inverse parameter-reduced dynamic models.
Also, as a blind search would result in an enormous computational effort,
 three different heuristics -$QR$, \textit{Backward Elimination} and
\textit{Forward Selection}- are proposed for the selection of the candidate parameter sets. The performance of the proposed methods is
tested on two examples of rigid multibody systems. As the high/low \textit{mobility} of bodies -defined as their ability to exhibit large/small excursions
in their respective motion space- plays an important role in the relevance of different dynamic parameters,
two multibody systems with very different characteristic mobility are analyzed.
Trajectory optimization is used to define sufficiently exciting characteristic data sets for the regressions. Model errors
and their algebraic
 operation count  are reported  for reduced models with a varying number of parameters obtained with the different
heuristics proposed.
Based on these results, we discuss the relevance of the proposed methods in the contexts of dynamic parameter estimation and computational complexity reduction in MSD.

The article is organized as follows: Section~\ref{sec::parameter-reduction} presents the methods and algorithms proposed in this article. First, in
Section~\ref{sec::base_parameter-reduction}, the classical base parameter determination \cite{Khalil1987} algorithm is proposed as a parameter reduction
method. A brief explanation of the algorithm is given, emphasizing this perspective. 
In Section \ref{sec::model_parameter-reduction}, the parameter reduction method
proposed in this paper is introduced as a natural extension of this view. 
%The normalized prediction
%error is defined and proposed as a performance measure for different parameter-reduced models.
The proposed heuristics are described in sections  \ref{sec::qr_algorithm},
	\ref{sec::backward_elimination} and
	\ref{sec::forward_addition}. In Section~\ref{sec::results},
	the algorithms introduced in this paper are applied to the selected examples. A detailed description of the analyzed systems and the results obtained are presented
	in Sections~\ref{sec::puma_robot} and \ref{sec::hexaglide_robot}. Section~\ref{sec::discussion} analyzes the results obtained,
	putting  the relevance of the methods proposed into perspective. Possible improvements and a more adequate
	normalized error criterion for  direct
	dynamic models are proposed and discussed. Finally, in section \ref{sec::conclusions}, the major outcomes of this work are highlighted.

% Moreover, once the reduced model regressors have been selected, the information of the eliminated regressors be can partially included in the reduced model, making this technique more accurate than simply eliminating the parameters.

%--------------------------------------------------
\section{Parameter Reduction Method}
 \label{sec::parameter-reduction}
% \subsection{Model reduction by parameter elimination}

Using a set of independent coordinates $\vz$, the Lagrange equations of a multibody system can be arranged as
\begin{equation}
 \mathbf{d}_{\vz}(\!\vz,\vdz,\vddz,\vphi)=\vtau_{\!\!\vz},
 \label{eq::lagrange_dynamic_model0}
\end{equation}
where $\mathbf{d}_{\vz}$ is the so-called \textit{inverse dynamics model} (IDM). The IDM makes it possible to obtain the vector of externally applied generalized forces $\vtau_{\!\!\vz}$, given 
the \textit{extended state} triad ($\vz$, $\vdz$, $\vddz$) and  the vector of dynamic parameters $\vphi$ of the system. In the context of rigid-body dynamics, symbolic implementations of recursive $O(n^3)$ formulations
constitute the state-of-the-art general purpose techniques for the efficient
computation of the IDM. The \textit{direct dynamics model} (DDM) is usually obtained by splitting out $\mathbf{d}_{\vz}$
 as
\begin{equation}
 \mathbf{d}_{\vz}(\!\vz,\vdz,\vddz,\vphi)=\M_{\!\vz\vz}(\vz,\vphi)\vddz + \vdelta_{\!\vz}(\vz,\vdz,\vphi)=\vtau_{\!\!\vz},
 \label{eq::lagrange_dynamic_model}
\end{equation}
where $\M_{\!\vz\vz}$ represents the mass matrix function and $\vdelta_{\!\vz}$ the generalized Coriolis,
centrifugal and constitutive forces function. Note that based on these DDM functions, given the state ($\vz$, $\vdz$) of the system  and the external generalized force vector $\vtau_{\!\!\vz}$,
the generalized accelerations can be obtained as follows:
\begin{equation}
\vddz=\M_{\!\vz\vz}^{-1}(\vz,\vphi) \left( \vtau_{\!\!\vz}-\vdelta_{\!\vz}(\vz,\vdz,\vphi) \right).
 \label{eq::direct_dynamics_problem}
\end{equation}
Regarding computational efficiency, state-of-the-art techniques to obtain these functions can be based on manipulations of
the symbolic expressions of $\mathbf{d}_{\vz}$, obtained using the aforementioned recursive formulations.

The ideas presented in this paper are more clearly explained making explicit the linearity of $\mathbf{d}_{\vz}$ with
respect to the dynamic parameters \cite{Atkeson1986}
\begin{equation}
 \mathbf{d}_{\vz}(\!\vz,\vdz,\vddz,\vphi)=\K_{\!\vz\vphi}(\vz,\vdz,\vddz)\vphi = \vtau_{\!\!\vz}.
 \label{eq::lp_dynamic_model}
\end{equation}
 This is a standard form of the IDM used in the literature on dynamic parameter estimation, where $\K_{\!\vz\vphi}$ is the single-sample observation matrix.
The goal of \textit{model simplification by parameter reduction} can be defined in this context
  as eliminating some  parameters from $\vphi$ and the corresponding columns of $\K_{\!\vz\vphi}$
so that vector $\vtau_{\!\!\vz}$ can be approximated as a linear combination of a subset of its columns as
\begin{equation}
 \K_{\!\vz\vphi_{\!R}}(\vz,\vdz,\vddz)\vphi_{\!R} \approx \K_{\!\vz\vphi}(\vz,\vdz,\vddz)\vphi = \vtau_{\!\!\vz},
 \label{eq::model_reduction}
\end{equation}
where $\vphi_{\!R}$ is the reduced parameter set and $\K_{\!\vz\vphi_{\!R}}$ the reduced single-sample observation matrix.
For IDM implementation purposes, it is preferable to eliminate the parameters directly from the recursively computed symbolic expressions
for $\mathbf{d}_{\vz}$, that is, approximating it as
\begin{equation}
  \mathbf{d}_{\vz}(\!\vz,\vdz,\vddz,\vphi) \approx \mathbf{d}_{\vz}(\!\vz,\vdz,\vddz,\vphi_{\!R}),
 \end{equation}
because in general this explicit linear form  $\K_{\!\vz\vphi_{\!R}}\vphi_{\!R}$ is computationally less efficient.
The computational cost of approximating $\vtau_{\!\!\vz}$ with the reduced model is obviously smaller.
As the functions $\M_{\!\vz\vz}$ and $\vdelta_{\!\vz}$  are also functions of $\vphi$, the DDM functions  $\M_{\!\vz\vz}$ and $\vdelta_{\!\vz}$
can be reduced using the same idea. For efficiency purposes, as  functions share common sub-expressions,
they are usually evaluated in a single function call. To illustrate this idea this function will be represented below as $[\M_{\!\vz\vz}|\vdelta_{\!\vz}](\vz,\vdz,\vphi)$.
% 
% Some authors \cite{Nguyen1989,Reboulet1991} have suggested to eliminate  parameters with small relative values of mass or inertia, 
% or those related to Coriolis terms, in the hope that their influence can be negligible. 

% However, a systematic criterion to decide whether or not a set of parameters can be eliminated without
% compromising the accuracy of the model has not been found in the multibody dynamics bibliography.

% To reduce the dynamic model complexity recursive formulations are frequently used to obtain the dynamic models.
% Most frequently, these formulations produce model functions in which the dynamic parameters $\vphi$
% are embedded:  $(\K_{\!\vz\vphi} \vphi)(\vz,\vdz,\vddz,\vphi)$ (for inverse dynamics), $[\M_{\!\vz\vz}|\vdelta_{\!\vz}](\vz,\vdz,\vphi)$\footnotemark (for direct dynamics).

So, for both IDM and DDM the computational complexity is reduced by symbolically simplifying the terms that are multiplied by the removed parameters in the
functions $ \mathbf{d}_{\vz}(\!\vz,\vdz,\vddz,\vphi)$ and $[\M_{\!\vz\vz}|\vdelta_{\!\vz}](\vz,\vdz,\vphi)$, respectively.
% in the different algebraic expressions.

%--------------------------------------------------
\subsection{Base Parameter Reduction} \label{sec::base_parameter-reduction}

In the context of dynamic parameter identification of robotic systems, the model is frequently re-parametrized
in terms of a smaller independent set of dynamic parameters.
One such parametrization is known as the base parameter set \cite{Gautier1991}, or minimum parameter set.
This parametrization, which leads to an exact model, can be regarded as a model reduction technique.

The parameter estimation problem is set up by ensuring that the dynamic equations shown in
Eq.~(\ref{eq::lp_dynamic_model}) are satified for a set of observations or estimation set,
${\cal E}=\left\{(^{i}\vz,^{i}\!\vdz,^{i}\!\vddz, ^{i}\!\vtau_{\!\!\vz}) | ~i=1,\ldots,n_{\cal E}\right\}$:
\begin{equation}
 \W({\cal E})\vphi=\begin{bmatrix}
     \K_{\!\vz\vphi}(^{1}\vz,^{1}\!\vdz,^{1}\!\vddz) \\
     \hdots \\
     \K_{\!\vz\vphi}(^{n}\vz,^{n}\!\vdz,^{n}\!\vddz)
    \end{bmatrix}\vphi~~
    =\vect{^{1}\!\vtau_{\!\!\vz}\\ \hdots \\ ^{n}\!\vtau_{\!\!\vz}}=\vchi({\cal E}),
    \label{eq::big_observation_matrix_definition}
\end{equation}
where $\W({\cal E})$ is the so-called observation matrix for the data set ${\cal E}$.

In general, no matter how ``exciting'' the estimation set is, $\W({\cal E})$ is rank deficient, meaning that there are linear dependencies
between its columns and, therefore, the set of parameters $\vphi$ is not independent.
$\W({\cal E})$ can be reordered as
\begin{equation}
 [\W_{\!\!\vphi_{\!R}},\W_{\!\!\vphi_{\!E}}]
\end{equation}
where matrix $\W_{\!\!\vphi_{\!R}}$ is made up of a chosen\footnote{there are several possible choices for this set of columns} set of ``independent" (or reduced) columns and matrix $\W_{\!\!\vphi_{\!E}}$ is formed by the remaining ``dependent" (or excluded) columns.
Parameters $\vphi$ can  be rearranged accordingly as
\begin{equation}
 \vphi=
 \begin{bmatrix}
  \vphi_{\!R}\\
  \vphi_{\!E}
 \end{bmatrix},
\end{equation}
where $\vphi_{\!R}$ are the set of ``independent" (or reduced) parameters associated with $\W_{\!\!\vphi_{\!R}}$
and $\vphi_{\!E}$  are the set of ``dependent" (or excluded) parameters associated with $\W_{\!\!\vphi_{\!E}}$.
The columns of $\W_{\!\!\vphi_{\!E}}$ can be expressed as linear combinations of the columns of matrix $\W_{\!\!\vphi_{\!R}}$ as:
% Therefore, it is possible to reorder $\vphi$ and $\W({\cal E})$  into ``independent" (or reduced) and ''dependent" (or excluded)
% parameters, $[\vphi_{\!R}^T,\vphi_{\!E}^T]^T$, and  columns, $[\W_{\!\!\vphi_{\!R}},\W_{\!\!\vphi_{\!E}}]$, so that
%
\begin{equation}
 \W_{\!\!\vphi_{\!E}}=\W_{\!\!\vphi_{\!R}} \vbeta_{\!\vphi_{\!R}}
 \label{eq::linear_combination_of_W}.
\end{equation}
In this expression, the columns of matrix $\vbeta_{\!\vphi_{\!R}}$ contain the coefficients of said linear combinations.
Using this relation, Eq.~(\ref{eq::big_observation_matrix_definition}) can be rewritten as
\begin{equation}
  [\W_{\!\!\vphi_{\!R}},\W_{\!\!\vphi_{\!E}}]\vect{\vphi_{\!R}\\ \vphi_{\!E}}=[\W_{\!\!\vphi_{\!R}},\W_{\!\!\vphi_{\!R}}\vbeta_{\!\vphi_{\!R}}]\vect{\vphi_{\!R}\\ \vphi_{\!E}}	
		=\W_{\!\!\vphi_{\!R}}(\vphi_{\!R}+\vbeta_{\!\vphi_{\!R}}\vphi_{\!E})=\W_{\!\!\vphi_{\!R}}\vphi_{\!\vphi_{\!R}}',
  \label{eq::perfect_model_reduction_W}
\end{equation}
leading to the base parameter identification problem
\begin{equation}
\W_{\!\!\vphi_{\!R}}({\cal E})\vphi_{\!R}'=\vchi({\cal E}),
\label{eq::perfect_model_reduction_Wb}
\end{equation}
where  
\begin{equation}
 \vphi_{\!R}'({\cal E})=\vphi_{\!R}+\vbeta_{\!\vphi_{\!R}}({\cal E}) ~\vphi_{\!E}
 \label{eq::base_parameter_expressions}
\end{equation}
is the so-called \emph{base} or \emph{minimum parameter set}. In this context,
the number of parameters in $\vphi_{\!R}$ is $n_{\vphi_{\!R}}=\mbox{rank}(\W({\cal E}))$. Note that this parametrization is dependent
on $\vphi_{\!R}$, the set of independent parameters chosen.
Also, for sufficiently exciting data sets (i.e. if ${\cal E}$ maximizes $\mbox{rank}(\W({\cal E}))$ and $\W_{\!\!\vphi_{\!R}}$ is well conditioned)
 $\vphi_{\!R}'$ and $\vbeta_{\!\vphi_{\!R}}$ become independent of ${\cal E}$. In particular $\vbeta_{\!\vphi_{\!R}}$ becomes dependent only on the geometric parameters of the system.

Accordingly, $\K_{\!\vz\vphi}$ can be split as $[\K_{\!\vz \vphi_{\!R}},\K_{\!\vz \vphi_{\!E}}]$.  Then the dynamic model can be expressed in terms of the base parameters $\vphi_{\!R}'$, as follows:
\begin{equation}
\K_{\!\vz \vphi_{\!R}} ~\vphi_{\!R}' = \vtau_{\!\!\vz}.
\label{eq::perfect_model_reduction}
\end{equation}
This equation can be regarded as a parameter-reduced model that reproduces
the original one exactly.
% This procedure applies as well when some constraint forces can not be removed from the model.
% To that end they should be incorporated into $\vtau_{\!\!\vz}$ as regular external generalized forces.
In \cite{Khalil1987} this reduced parametrization is applied to the IDM of
some serial manipulators to reduce their operation count without compromising their accuracy.

% If the model reduction is performed by simply eliminating $\W_2\vphi_2$ from Equation (\ref{eq::perfect_model_reduction_W}), the remaining parameters will represent the so called \emph{minimum parameters} \cite{Shome1998}. In this situation, while the model observation matrix will be the same as in the base parameters approach, the physical sense of the parameters will be lost. Let us therefore note that the reduced model in terms of base parameters actually includes the information of the parameters that multiply the eliminated colums ($\W_2$). Advantage of this property will be taken when the more general model reduction method is described.

% The kinematic couplings of the joints in a multibody system are the responsible for the linear combinations (rank deficiency) of matrix $\W$. In fact, the base parameter expressions of Equation (\ref{eq::base_parameter_expressions}) can be obtained symbolically \cite{Khalil1995,Ros2012} with the sole information of the kinematics of the system.  However, for a number of reasons, some \emph{nearly} linear combinations of the columns of $\W$ can exist, and treating those columns as if they where \emph{perfect} linear combinations of the other, the model can be further reduced.
% Unlike in Equation (\ref{eq::perfect_model_reduction_W}), in this model reduction approach a simplification of the model will be done. 

%--------------------------------------------------
\subsection{Model parameter reduction}	\label{sec::model_parameter-reduction}

The results in the context of parameter estimation of multibody systems show that frequently not all the
base or minimum parameters can be estimated with "sufficient" accuracy. This may be related to the
fact that for a given estimation data set, ${\cal E}$, the columns of matrix $\W({\cal E})$ can be expressed with ``sufficient" precision as  linear combinations of
a subset  $\W_{\!\!\vphi_{\!R}}$ of $n_{\vphi_{\!R}}$, $n_{\vphi_{\!R}}<\mbox{rank}(\W({\cal E}))$, independent columns of $\W({\cal E})$. It can
be said that the characteristic data set does not sufficiently excite certain system dynamics.

Let
$\vphi_{\!R}$ be the parameters associated with the chosen independent column subset $\W_{\!\!\vphi_{\!R}}$. The associated reduced model  will
be referred to as the $\vphi_{\!R}$-parameter-reduced model, or $\vphi_{\!R}$ model for short.

Let $\W_{\!\!\vphi_{\!E}}$ be the set of removed columns and $\vphi_{\!E}$ their associated parameters. Proceeding as
in the previous section, the columns of $\W({\cal E})$ and the set of parameters $\vphi$
can be reordered as  $\W({\cal E})=[\W_{\!\!\vphi_{\!R}},\W_{\!\!\vphi_{\!E}}]$ and $\vphi=[\vphi_{\!R}^T,\vphi_{\!E}^T]^T$.
 $\W_{\!\!\vphi_{\!E}}$ can be approximated by its projection onto the column space of $\W_{\!\!\vphi_{\!R}}$ as:
\begin{equation}
 \W_{\!\!\vphi_{\!E}} \approx  \W_{\!\!\vphi_{\!R}}{\W_{\!\!\vphi_{\!R}}^+}\W_{\!\!\vphi_{\!E}}=\W_{\!\!\vphi_{\!R}} \vbeta_{\!\vphi_{\!R}},
 \label{eq::reduced_W_E}
\end{equation}
where 
\begin{equation}
\vbeta_{\!\vphi_{\!R}}={\W_{\!\!\vphi_{\!R}}^+}\W_{\!\!\vphi_{\!E}}.
\label{eq::reduced_beta}
\end{equation}
Based on the preceding two equations, $\W({\cal E})\vphi$ can be approximated as:
\begin{equation}
\begin{split}
   [\W_{\!\!\vphi_{\!R}},\W_{\!\!\vphi_{\!E}}]\vect{\vphi_{\!R}\\ \vphi_{\!E}}\approx[\W_{\!\!\vphi_{\!R}},\W_{\!\!\vphi_{\!R}}\vbeta_{\!\vphi_{\!R}}]\vect{\vphi_{\!R}\\ \vphi_{\!E}}
		=\W_{\!\!\vphi_{\!R}}(\vphi_{\!R}+\vbeta_{\!\vphi_{\!R}}\vphi_{\!E})=\W_{\!\!\vphi_{\!R}}\vphi_{\!R}',
  \label{eq::approximate_model_reduction_from_full_model}
\end{split}
\end{equation}
leading to the approximate estimation problem:
\begin{equation}
 \W_{\!\!\vphi_{\!R}}({\cal E})\vphi_{\!R}'\approx\vchi({\cal E}).
\end{equation}
The previous relations show that the least square solution to this estimation problem is:
\begin{equation}
\vphi_{\!R}'({\cal E})={\W_{\!\!\vphi_{\!R}}^+}({\cal E})\,\vchi({\cal E})=\vphi_{\!R}+\vbeta_{\!\vphi_{\!R}}({\cal E})\vphi_{\!E},
\label{eq::generalized_base_parameter}
\end{equation}
where $\vbeta_{\!\vphi_{\!R}}({\cal E})$ can be obtained using Eq.~(\ref{eq::reduced_beta}).

Eq.~(\ref{eq::generalized_base_parameter}) can be seen as a generalization of the \emph{base parameter}
concept  when the number of parameters used is smaller than the
rank of $\W({\cal E})$, $n_{\vphi_{\!R}}<\mbox{rank}(\W({\cal E}))$, that is, a generalization of the concept of base parameters to the context of
approximate parameter-reduced models. We use the term ``approximate base parameter set'' or ``approximate minimum parameter set'' to distinguish it from the
exact base parametrization described in the previous section.  This equation can be used
to compute the values of parameters $\vphi_{\!R}'({\cal E})$ from the original model values $\vphi$. Additionally,
this equation provides information on the influence that the different parameters have on the dynamics. Note that
this information is not contained in the bare numerical values of $\vphi_{\!R}'({\cal E})$.

The parameter-reduced model selection problem can be formulated as follows:
\begin{center}
\textit{``Given a dynamic model with a known parameter set determine a minimal set of parameters of the model that
approximates the characteristic model data with the desired accuracy.''}
\end{center}
This leads to the questions of defining: 1) the characteristic estimation data set, and 2) the error measure.
\begin{itemize}
 \item[1)] The characteristic estimation data set is system and application dependent, and should characterize the system's entire dynamic range
in the considered application.
This set, ${\cal E}=\left\{(^{i}\vz,^{i}\!\vdz,^{i}\!\vddz, ^{i}\!\vtau_{\!\!\vz}) | ~i=1,\ldots,n_{\cal E}\right\}$,
can be obtained by sampling data from full model simulations.

For example, in the case of fully-actuated multibody systems,
trajectories are parametrized for a set of independent coordinates, and an optimization criterion is used to determine the values of their parameters.
A comparison of some of the classic criteria for trajectory optimization can be found in \cite{Sun2008}. Trajectory parametrizations based on harmonic series  \cite{Calafiore98,Swevers1996,Park2006}
and on polynomials \cite{Armstrong1989,Gautier1992a} have been proposed in the literature.
For underactuated systems, dynamic simulations must be performed for a set of representative working simulations. 
The validation data set, ${\cal V}=\left\{(^{i}\vz,^{i}\!\vdz,^{i}\!\vddz, ^{i}\!\vtau_{\!\!\vz}) | ~i=1,\ldots,n_{\cal V}\right\}$, is obtained using the same procedure.
%The validation data is used in this context to ascertain the quality of the estimation data in a statistical sense. 
% The estimation data can be said to be characteristic enough when the validation data is reproduced accurately
% by the reduced model. This, in turn, speaks about the reduced model quality.

\item[2)] For a given $\vphi_{\!R}$ parameter-reduced  model, the following
\textit{normalized error criterion} or \textit{prediction error measure} for the {IDM} is proposed:
\begin{equation}
 \epsilon_{\vtau_{\!\!\vz}}(\vphi_{\!R},{\cal E})=\frac{\norm{\sigmamenosunmedio(\vchi({\cal E})-\W_{\!\!\vphi_{\!R}}({\cal E})\,\vphi_{\!R}'({\cal E}))}}{\norm{\sigmamenosunmedio\vchi ({\cal E})}}.
 \label{eq::reduction_error_tau_a} 
\end{equation}
In this expression ${\W_{\!\!\vphi_{\!R}}}({\cal E})$ is the observation matrix of the
$\vphi_{\!R}$-parameter-reduced  model determined using the data set ${\cal E}$. 
Parameters $\vphi_{\!R}'({\cal E})$ are the numerical values for the generalized base parameters, defined in Eq.~\ref{eq::generalized_base_parameter}, for the $\vphi_R$ model using the data set ${\cal E}$.
The weighting matrix $\sigmaunmedio$ is defined as:
\begin{equation}
\sigmaunmedio=
diag(diag(\mbox{nom}(\vtau_{\!\!\vz})), \ldots, diag(\mbox{nom}(\vtau_{\!\!\vz}))).
\end{equation}
where $\mbox{nom}(\vtau_{\!\!\vz})$ represents a vector of characteristic values for the elements of vector $\vtau_{\!\!\vz}$. In general,
characteristic values based on the estimation data can be used (e.g. standard deviation), although they may be established by other means (e.g. nominal torque value of an actuator).

It is important to note that for function $\epsilon_{\vtau_{\!\!\vz}}(\vphi_{\!R},{\cal E})$ the argument $\vphi_{\!R}$ refers to the reduced set of parameters and not to their numerical values. 
Obviously, the computation of this function requires the determination of the numerical values $\vphi_{\!R}'({\cal E})$.

Indexes to characterize the relevance of the contribution of different dynamic forces (Coriolis, centripetal, etc.) to the IDM have been proposed \cite{Wiens2002},
but these are not well suited to the parameter reduction problem on which this work focuses.

In a more programmatic style, the error criteria can be expressed equivalently as:
\begin{equation}
 \epsilon_{\vtau_{\!\!\vz}}(\vphi_{\!R},{\cal E})=\frac{\norm{~\mbox{col}(~\{\mbox{diag}(~\mbox{nom}({\vtau_{\!\!\vz}}))^{-1}~(^{i}\mathbf{d}_{\vz}- {^{i}\mathbf{d}_{\vz}(\vphi_{\!R}'}({\cal E}))~|~ \forall i\in{\cal E}\}~)~}}{\norm{~\mbox{col}(~\{\mbox{diag}(\mbox{nom}({\vtau_{\!\!\vz}}))^{-1}~^{i}\mathbf{d}_{\vz} ~|~ \forall i\in{\cal E}\} ~)~}}.
 \label{eq::reduction_error_tau} 
\end{equation}
where $\mbox{col}(\cal S)$ arranges the elements of  $\cal S$ into a column, and $^{i}\mathbf{d}_{\vz}$ and ${^{i}\mathbf{d}_{\vz}}(\vphi_{\!R}'({\cal E}))$ are external
forces for the $i$-th data sample predicted by the full and $\vphi_{\!R}$ parameter-reduced IDM, respectively. Note that, for short, ${^{i}\mathbf{d}_{\vz}}(\vphi_{\!R}'({\cal E}))$
is interpreted as ${\mathbf{d}_{\vz}}(^{i}\vz,^{i}\vdz,^{i}\vddz, [\vphi_{\!R}'({\cal E}),\mathbf{0}])$ where ${\mathbf{d}_{\vz}}(^{i}\vz,^{i}\vdz,^{i}\vddz,[\vphi_{\!R},\vphi_{\!E}])={\mathbf{d}_{\vz}}(^{i}\vz,^{i}\vdz,^{i}\vddz,\vphi)$,
and $^{i}\mathbf{d}_{\vz}$ as ${\mathbf{d}_{\vz}}(^{i}\vz,^{i}\vdz,^{i}\vddz,\vphi)$. $\vphi$ are the actual parameter values and $\vphi_{\!R}'({\cal E})$ are the generalized base parameters defined in Eq.~\ref{eq::generalized_base_parameter}.

In Eq.~(\ref{eq::reduction_error_tau_a}) and (\ref{eq::reduction_error_tau}), the proposed norm can be defined in different ways. The most interesting norms in this
context are probably the Euclidean norm (chosen in the examples in this paper), the $1$-norm and the infinity norm.

In this paper, this error measure, $\epsilon_{\vtau_{\!\!\vz}}(\vphi_{\!R},{\cal E})$,  is used as the fitness criterion for the choice of parameter-reduced models and for their validation.
For example, if the number of parameters, $n_{\vphi_{\!R}}$, of the desired $\vphi_{\!R}$ model is fixed, the best model can be determined as
\begin{equation}
\vphi_R=
\displaystyle \argmin_{
\substack{
\vphi_{\!R} ~|~
n_{\vphi_{\!R}}=\textnormal{card}(\vphi_{\!R})
       }
} \epsilon_{\vtau_{\!\!\vz}}(\vphi_{\!R},{\cal E}),
\end{equation}
where $\mbox{card}(\vphi_{\!R})$ is the cardinality of the set of parameters $\vphi_{\!R}$. As before, note that the argument $\vphi_{\!R}$
refers to the reduced set of parameters chosen and not to their numerical values.
For the desired error level, $\epsilon_{\vtau_{\!\!\vz}}(\vphi_{\!R},{\cal E})$, the reduced model is considered
acceptable if validation and estimation data give similar
error levels $\epsilon_{\vtau_{\!\!\vz}}(\vphi_{\!R},{\cal V})\approx\epsilon_{\vtau_{\!\!\vz}}(\vphi_{\!R},{\cal E})$.

Analogously, the \textit{normalized error criterion} or \textit{normalized prediction error} for the \textit{DDM} can be defined as
\begin{equation}
 \epsilon_{\vddz}(\vphi_{\!R},{\cal E})=\frac{\norm{~\mbox{col}(~\{\mbox{diag}(~\mbox{nom}({\vddz}))^{-1}~(^{i}\vddz- {^{i}\vddz}(\vphi_{\!R}'({\cal E})))~|~ \forall i\in{\cal E}\}~)~}}{\norm{~\mbox{col}(~\{ \mbox{diag}(\mbox{nom}({\vddz}))^{-1}~^{i}\vddz~|~ \forall i\in{\cal E} \}~)~}},
 \label{eq::reduction_error_accel} 
\end{equation}
where $^{i}\vddz$ and ${^{i}\vddz}(\vphi_{\!R}'({\cal E}))$ are the accelerations for the $i$-th data sample predicted by the full and $\vphi_{\!R}$ parameter-reduced DDM, respectively.
Although the model selection performed in this article is not  based on this error, to enrich the discussion, values of this error measure will be computed for the parameter-reduced models
 obtained based on $\epsilon_{\vtau_{\!\!\vz}}(\vphi_{\!R},{\cal E})$.

\end{itemize}

The problem with this approach is that the number of possible parameter sets
to be tried in order to find the best parameter-reduced  model with $\mbox{card}(\vphi_{\!R})=n_{\vphi_{\!R}}$ is huge:
\begin{equation}
\left(\begin{matrix}n_{\vphi}\\ n_{\vphi_{\!R}}\end{matrix}\right) = \frac{n_{\vphi}!}{n_{\vphi_{\!R}}!(n_{\vphi}-n_{\vphi_{\!R}})!}.
\end{equation}
If  all parameter sets with an arbitrary number of parameters $n_{\vphi_{\!R}}=1,\ldots,n_{\!\vphi}$ are tried,
 a much larger number of candidate sets, $2^{n_{\!\vphi}}$, is obtained.  In general, this excludes a trial-and-error approach to the parameter reduction problem.
Therefore, algorithms that find a compromise between the
computational cost of the search and the quality of the reduced model are required. To that end,
three candidate model selection heuristics are proposed: $QR$ selection in Section~\ref{sec::qr_algorithm}, \textit{Backward Elimination} in Section~\ref{sec::backward_elimination}, and \textit{Forward Selection}  in Section~\ref{sec::backward_elimination}.

%--------------------------------------------------
\subsection{$QR$ decomposition-based heuristic}	\label{sec::qr_algorithm}

% This decomposition implicitly introduces a relevance  order in the parameter set, $\vphi$. More/less relevant parameters are those  corresponding to bigger/smaller diagonal elements in  matrix $\R$. Therefore it seems logical to try this relevance as a possible elimination heuristics: less relevant parameters are removed first.

A prototypical $QR$ decomposition with column pivoting of the observation matrix $\W({\cal E})$ takes the following form:
\begin{equation}
 \W \E = \Q \R,
\end{equation}
where matrix $\W \E$ is a column permutation of $\W({\cal E})$,  $\Q$ is an orthonormal matrix, and  $\R$ is an upper triangular
matrix with diagonal elements of decreasing magnitude.

Let $r=n_{\vphi_{\!R}}\le \mbox{rank}(\W({\cal E}))$.  $\W\E$ is split as
\begin{equation}
[\W_{\!\!\vphi_{\!R}},\W_{\!\!\vphi_{\!E}}]=[\Q_R,\Q_E]
	\begin{bmatrix}
    \R_{rr} & \R_{re} \\ \vzero  & \R_{ee}
    \end{bmatrix},
\end{equation}
where $\W_{\!\!\vphi_{\!R}}$ contains the first $r$ columns of $\W\E$, and $\R_{rr}$ is an $r\times r$ regular matrix.
The  reduced model is defined by the parameter set $\vphi_{\!R}$ associated with the
columns of $\W_{\!\!\vphi_{\!R}}$. From the previous equation it follows that
\begin{equation}
 \W_{\!\!\vphi_{\!R}}=\Q_R\R_{rr} \quad \text{and} \quad \W_{\!\!\vphi_{\!E}}=\Q_R\R_{re} +\Q_E \R_{ee}.
\end{equation}
Based on Eq.~(\ref{eq::reduced_W_E})
 matrix $\W_{\!\!\vphi_{\!E}}$  can be approximated as:
\begin{equation}
 \W_{\!\!\vphi_{\!E}}\approx\W_{\!\!\vphi_{\!R}}{\W_{\!\!\vphi_{\!R}}^+}\W_{\!\!\vphi_{\!E}}=\Q_R\R_{re}=\W_{\!\!\vphi_{\!R}} \vbeta_{\!\vphi_{\!R}},
 \label{eq::approx_WE_QR}
\end{equation}
where
\begin{equation}
 \vbeta_{\!\vphi_{\!R}}={\W_{\!\!\vphi_{\!R}}^+}\W_{\!\!\vphi_{\!E}}={\R_{rr}\!\!\!\!}^{-1}\R_{re}.
\end{equation}
Eq.~(\ref{eq::approx_WE_QR}) is a very good approximation of $\W_{\!\!\vphi_{\!E}}$ for
a parameter set of $r$ elements. Note that $\R_{ee}$ is chosen by the $QR$ algorithm
to have the smallest diagonal elements and, therefore,  the error in the approximation of $\W_{\!\!\vphi_{\!E}}$, $\Q_E \R_{ee}$,
is nearly  as small as possible. The ideas presented here are partly
inspired by the $QR$ decomposition-based base parameter determination algorithm
described in \cite{Gautier1991}, although the ideas presented there focus on the case of $r=n_{\vphi_{\!R}}=\mbox{rank}(\W({\cal E}))$.
A subset selection algorithm with some resemblance to the one  presented here, referred to as \textit{Orthogonal Least Squares}, is described in \cite{Nelles2001}.

The normalized prediction error $\epsilon_{\vtau_{\!\!\vz}}(\vphi_{\!R},{\cal E})$ corresponding to the parameters associated with the first $r$
 columns of $\W\E$  can be computed using Eq.~(\ref{eq::reduction_error_tau_a}).
Given a maximum acceptable error tolerance, $tol$, and starting
with $n_{\vphi_{\!R}}=\mbox{rank}(\W({\cal E}))$, we decrease $n_{\vphi_{\!R}}$ while  $\epsilon_{\vtau_{\!\!\vz}}(\vphi_{\!R},{\cal E}) < tol$. Then, the simplest
parametrization based on the $QR$ heuristics is that corresponding
to the parameters associated with the first $n_{\vphi_{\!R}}$ rows of $\W\E$.

A possible criticism of this algorithm is that the parameter order obtained
is only dependent on $\W({\cal E})$, but not on $\vchi({\cal E})$.
The results presented in Section~\ref{sec::results} will prove that the  heuristics proposed in the next
two sections perform much better than this one.

%--------------------------------------------------
\subsection{\textit{Backward Elimination} heuristic} \label{sec::backward_elimination}

The \textit{Backward Elimination} (BE) algorithm is an iterative algorithm. The algorithm starts by using the full set
of parameters for the current parameter-reduced model, $\vphi_{\!R}$$=$$\vphi$.
At each iteration, the parameter with the least significant contribution to the normalized error,
\begin{equation}
 \varphi_e= \argmin_{\substack{\varphi_i ~|~ \varphi_i \in \vphi_{\!R}}} \epsilon_{\vtau_{\!\!\vz}}(\vphi_{\!R} \smallsetminus \varphi_i,{\cal E}),
\end{equation}
 is eliminated from the current model parameter set, $\vphi_{\!R}=\vphi_{\!R} \smallsetminus \varphi_e$, 
until the normalized error criterion is greater than the chosen tolerance, $\epsilon_{\vtau_{\!\!\vz}}(\vphi_{\!R} , {\cal E} ) > tol $.
This is shown in pseudocode in Algorithm~\ref{alg::backward_elimination} using a more formal algorithmic style.
\begin{algorithm}
\caption{Backward Elimination (BE)}\label{alg::backward_elimination}
\normalsize
\begin{algorithmic}[]
\State $\vphi_{\!R}=\vphi$
\Repeat
\State $\varphi_e=
\displaystyle \argmin_{\substack{\varphi_i ~|~ \varphi_i \in \vphi_{\!R}}} \epsilon_{\vtau_{\!\!\vz}}(\vphi_{\!R} \smallsetminus \varphi_i,{\cal E})$
\State $\vphi_{\!R}=\vphi_{\!R} \smallsetminus \varphi_e$
\Until{$\epsilon_{\vtau_{\!\!\vz}}(\vphi_{\!R},{\cal E}) > tol $}
\end{algorithmic}
\end{algorithm}

Basically, the parameters that make a \textit{smaller} contribution to the modeling error are \textit{removed} one by one.
The algorithm would be optimal if the contributions to the error of the different parameters, $\vphi$,
were independent. The issue is that these contributions are not generally independent, so \textit{eliminating} a
given parameter has an effect on the weight that the remaining parameters have for the resulting reduced model.
Therefore, the algorithm is likely to perform well but not optimally.

An important feature of this algorithm is that, in addition to $\W({\cal E})$, the vector $\vchi({\cal E})$ has an effect on the order in which the parameters
are removed. This information is not considered
by the previously introduced $QR$ heuristic, so this may be a plus.

%--------------------------------------------------
\subsection{\textit{Forward Selection} heuristic}	\label{sec::forward_addition}

The iterative \textit{Forward Selection} (FS) algorithm is, in a way, the reverse of the previous one. It starts
with an empty  parameter set for the current parameter-reduced model, $\vphi_{\!R}$$=$$\varnothing$.
At each iteration, the parameter with the most significant contribution to the normalized error, 
\begin{equation}
 \varphi_s=
 \argmin_{\substack{\varphi_i\\ \varphi_i \notin \vphi_{\!R}}} \epsilon_{\vtau_{\!\!\vz}}(\vphi_{\!R} \cup \varphi_i,{\cal E}),
\end{equation}
 is added to the current model parameter set, $\vphi_{\!R}=\vphi_{\!R} \cup \varphi_s$,
until the required tolerance objective is met, $\epsilon_{\vtau_{\!\!\vz}}(\vphi_{\!R}, {\cal E} ) < tol $.
This is shown in pseudocode in Algorithm~\ref{alg::forward_addition} using a more formal algorithmic style.
\begin{algorithm}
\caption{Forward Selection (FS)}\label{alg::forward_addition}
\normalsize
\begin{algorithmic}[]
\State $\vphi_{\!R}=\varnothing$
\Repeat
\State $\varphi_s=
\displaystyle \argmin_{\substack{\varphi_i ~|~ \varphi_i \notin \vphi_{\!R}}} \epsilon_{\vtau_{\!\!\vz}}(\vphi_{\!R} \cup \varphi_i,{\cal E})$
\State $\vphi_{\!R}=\vphi_{\!R} \cup \varphi_s$
\Until{$\epsilon_{\vtau_{\!\!\vz}}(\vphi_{\!R} , {\cal E} ) < tol $}
\end{algorithmic}
\end{algorithm}

In essence, the parameters that make a \textit{larger} contribution to the modeling error are \textit{introduced} one by one.
As before, the algorithm would be optimal if the contributions to the error of the different parameters, $\vphi$,
 were independent. The issue is that these contributions are not  generally independent, so \textit{adding} a
given parameter has an effect on the weight that the parameters already introduced have for the resulting reduced model.
Therefore, the algorithm is likely to perform well but not optimally.

Later in this paper, it will be seen that FS performs slightly better than the
BE heuristic: at the beginning of BE, the model can be expressed exactly in terms of the base parameter set.
As this set is not unique,
there are several possible parameter removals that have no effect whatsoever on the error. Therefore, 
there are no criteria to differentiate among possible parameter eliminations but, still,
the choice has an effect on the performance of the resulting parameter-reduced model. The FS algorithm
never has to face the selection of noncontributing parameters, as these are never added.

%--------------------------------------------------
\subsection{Other Forward and Backward heuristics}	\label{sec::other_forward_backward_heuristics}
The algorithms presented in Sections \ref{sec::backward_elimination} and \ref{sec::forward_addition} are analogous to the \emph{forward selection} and
\emph{backward elimination} algorithms typically found in the context of model selection \cite{Miller1990,Nelles2001}.
The literature suggests the possibility of using more complex \textit{step-wise} \cite{Nelles2001} iterations in the
forward/backward selection/elimination heuristics. Typically, one or more search iterations are introduced in addition to the initial one.

For example (adapted from \cite{Miller1990}), a possible forward selection algorithm with an additional search iteration would read:
let $\vphi_{\!R}$ be the set of parameters at the start of a given iteration, and let $\varphi_l$ be the parameter added
to this set in the previous iteration. The algorithm starts with an empty parameter set for the current parameter-reduced model, $\vphi_{\!R}$$=$$\varnothing$, and an empty  variable as the one added in the previous iteration, $\varphi_l=\varnothing$. 
At each iteration of the forward selection, as before, the parameter with the most significant contribution to the normalized error,
\begin{displaymath}
 \varphi_s=
\displaystyle \argmin_{\substack{\varphi_i ~|~ \varphi_i \notin \vphi_{\!R}}} \epsilon_{\vtau_{\!\!\vz}}(\vphi_{\!R} \cup \varphi_i,{\cal E}),
\end{displaymath}
is added to the current  model, $\vphi_{\!R}=\vphi_{\!R} \cup \varphi_s$. Then,
the parameter added in the previous iteration, $\varphi_l$, is removed, $\vphi_{\!R}=\vphi_{\!R} \smallsetminus  \varphi_l$.
 Next, the parameter with the most significant contribution to the normalized error is searched anew,
\begin{displaymath}
 \varphi_s=
\displaystyle \argmin_{\substack{\varphi_i ~|~ \varphi_i \notin \vphi_{\!R}}} \epsilon_{\vtau_{\!\!\vz}}(\vphi_{\!R} \cup \varphi_i,{\cal E}),
\end{displaymath}
 and added to the set, $\vphi_{\!R}=\vphi_{\!R} \cup \varphi_s$, until
the error tolerance is met, {$\epsilon_{\vtau_{\!\!\vz}}(\vphi_{\!R}, {\cal E} ) < tol $}.
This is shown more formally in the pseudocode in Algorithm~\ref{alg::forward_addition_2}.

\begin{algorithm}
\caption{Forward Selection with two iterations}\label{alg::forward_addition_2}
\normalsize
 \begin{algorithmic}[]
 \State $\vphi_{\!R}=\varnothing$\;
 \State $\varphi_l=\varnothing$\;
\Repeat
\State $\varphi_s=
\displaystyle \argmin_{\substack{\varphi_i ~|~ \varphi_i \notin \vphi_{\!R}}} \epsilon_{\vtau_{\!\!\vz}}(\vphi_{\!R} \cup \varphi_i,{\cal E})$
\State $\vphi_{\!R}=\vphi_{\!R} \cup \varphi_s$
\State $\vphi_{\!R}=\vphi_{\!R} \smallsetminus  \varphi_l$
\State $\varphi_s=
\displaystyle \argmin_{\substack{\varphi_i ~|~ \varphi_i \notin \vphi_{\!R}}} \epsilon_{\vtau_{\!\!\vz}}(\vphi_{\!R} \cup \varphi_i,{\cal E})$
\State $\varphi_l=\varphi_s$
\Until{$\epsilon_{\vtau_{\!\!\vz}}(\vphi_{\!R}, {\cal E} ) < tol $}
\end{algorithmic}
\end{algorithm}

This algorithm is not going
be tested on the application examples analyzed in the next section. It has just been presented here to give some
perspective to the ideas presented in this paper.

%--------------------------------------------------
\section{Examples and Simulations}
\label{sec::results}

In this article, high/low \textit{mobility} of a body in a multibody system is defined as the possibility of the body to exhibit large/small excursions
in its relative motion space with respect to other bodies and/or with respect to the inertial frame.
It plays an important role in the relevance of different dynamic parameters.
For example, the ideas presented in \cite{Ros2012} show that for each d.o.f lost between two arbitrary
bodies, or between an arbitrary body and the inertial frame, the number of base parameters of the system can be reduced.
Thus, if some relative motion
between two bodies or between a body and the ground has a small amplitude, the  relevance of some parameters in the dynamics
might be small. Therefore, in these cases, it is expected that precise parameter-reduced models requiring a
relatively small number of parameters will be obtained.

This is why we propose to test the different parameter reduction strategies on two different multibody systems with different -high and low- characteristic mobility.
The results for the selected high-mobility system are presented in Section~\ref{sec::puma_robot} and the results for the low-mobility
system are given in Section~\ref{sec::hexaglide_robot}.

First, the system description, the modeling approach used and the procedure followed to obtain
the characteristic estimation, ${\cal E}$,  and validation, ${\cal V}$, data sets are described in detail. Then,
the results of  applying the proposed heuristics  to obtain parameter-reduced
models with an arbitrary number of parameters, $n_{\vphi_{\!R}}=1,\ldots,n_\vphi$, are presented
 in a way that is appropriate for the discussion. As stated before,  the normalized error for the inverse dynamics model, $\epsilon_{\vtau_{\!\!\vz}}(\vphi_{\!R}, {\cal E} )$,
 is the performance measure used to obtain every parameter-reduced model.

%--------------------------------------------------
\subsection{High-mobility multibody example} \label{sec::puma_robot}

\begin{figure}[ht]
  \begin{center}
    \includegraphics[width=10.32cm]{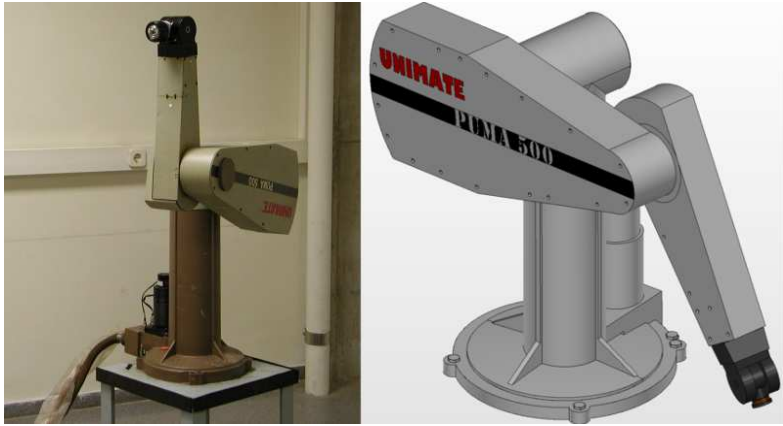}
    \caption{High-mobility example: left) Actual system, right) CAD model.} 
    \label{fig:Puma_foto_3d} 
  \end{center}
\end{figure}

The six \textit{d.o.f.} PUMA$560$ serial manipulator, shown in Fig.~\ref{fig:Puma_foto_3d}, is the high-mobility multibody system example (HME) selected.
The \textit{modified Denavit-Hartenberg} \cite{Khalil2002} kinematic parameters and coordinates and the inertia parameter
set used  to define the model, taken from reference \cite{Benimeli2006}, are shown in Tables \ref{tab::puma_denavit_hartenberg} and \ref{tab::puma_inertia_parameters}, respectively. 

The \textit{Principle of Virtual Power}  has been used to obtain the dynamic equations. As joint/relative coordinates are used for
the modeling, no constraint equations and no Lagrange multipliers are required.
The symbolic library in \cite{Ros2007} has been used to obtain the  IDM and DDM functions, $\mathbf{d}_{\vz}(\vz,\vdz,\vddz,\vphi_{\!R})$ and $[\M_{\!\vz\vz}|\vdelta_{\!\vz}](\vz,\vdz,\vphi_{\!R})$, respectively.
This library strongly minimizes the number of algebraic operations using algorithms compatible with  $O(n^3)$ recursive formulations, with a performance that is on a par with state-of-the-art formulations.
In this way, the reported simplifications are performed on already quasi-optimal models. This ensures fair reporting of the computational complexity of the reduced models, which is given in terms of the number of
operations. 

The nominal torques of the actuators, $\mbox{nom}({\vtau}_{\!\!\vz})=\left[350,300,125,8,3,1\right]\mbox{{N}{m}}$, are used to
define the normalized prediction error for the IDM, $\epsilon_{\vtau_{\!\!\vz}}(\vphi_{\!R},{\cal E})$.
\begin{table}[ht]
\caption{Modified Denavit-Hartenberg parameters of the PUMA$560$.}
\label{tab::puma_denavit_hartenberg}
\begin{center}
\footnotesize
\begin{tabular}{ccccc}
      \hline
{Body}	&	${a_i}$&	${\alpha_i}$	& ${d_i}$ 	&	${\theta_i}$ 	\\
%  \hline
{Units}	&	m&	rad	& m 	&	rad 	\\
\hline
%\hline
1		&	0		&	0			&	0		&	$z_1$	\\
% \hline
2		&	0		&	$-\pi/2$		&	0		&	$z_2$	\\
% \hline
3		&	0.4318		&	0			&	-0.1491		&	$z_3$	\\
% \hline
4		&	-0.0203	&	$\pi/2$			&	-0.4318		&	$z_4$	\\
% \hline
5		&	0		&	$-\pi/2$		&	0		&	$z_5$	\\
% \hline
6		&	0		&	$\pi/2$			&	0		&	$z_6$	\\
\hline
\end{tabular}
\end{center}
\end{table}

\begin{table}[!ht]
\caption{Dynamic parameters for PUMA$560$.}
\label{tab::puma_inertia_parameters}
\label{tab::puma_dyn_param}
\begin{center}
\footnotesize
\begin{tabular}{ccccccccccc}
\hline
{Body} &
$m$ & $d_x$ & $d_y$ & $d_z$ & $I_{xx}$ & $I_{xy}$ & $I_{xz}$ & $I_{yy}$ & $I_{yz}$ & $I_{zz}$ \\
% \hline
{Unit} & kg & kg~m & kg~m & kg~m & kg~m$^2$ & kg~m$^2$ & kg~m$^2$ & kg~m$^2$ & kg~m$^2$ & kg~m$^2$ \\
% \hline
\hline
$1$ &  $10.52$ &  $0.0$ &  $-0.568$ &  $0.0$ &  $1.643$ &  $0.0$ &  $0.0$ &  $0.509$ &  $0.0$ &  $1.643$ \\
% \hline
$2$ &  $15.78$ &  $2.206$ &  $0.2$ &  $2.353$ &  $0.841$ &  $0.2$ &  $-0.329$ &  $8.738$ &  $0.4$ &  $8.576$ \\
% \hline
$3$ &  $8.767$ &  $-0.003$ &  $-1.727$ &  $0.0$ &  $3.717$ &  $-0.001$ &  $0.002$ &  $0.301$ &  $0.002$ &  $3.717$ \\
% \hline
$4$ &  $1.052$ &  $0.03$ &  $0.06$ &  $-0.060$ &  $0.184$ &  $0.000$ &  $0.0$ &  $0.184$ &  $0.000$ &  $0.127$ \\
% \hline
$5$ &  $1.052$ &  $0.004$ &  $-0.007$ &  $0.005$ &  $0.074$ &  $0.000$ &  $0.000$ &  $0.074$ &  $0.000$ &  $0.127$ \\
% \hline
$6$ &  $0.351$ &  $0.01$ &  $0.02$ &  $0.013$ &  $0.008$ &  $0.000$ &  $0.002$ &  $0.008$ &  $0.000$ &  $0.014$ \\
\hline
\end{tabular}
\end{center}
\end{table}

\subsubsection*{Estimation trajectories}

The \emph{condition number} of the observation matrix $\kappa(\W({\cal E}))$\footnote{defined as $\kappa(\W({\cal E}))=\sigma_1/\sigma_r=\norm{\W}_2 \norm{\W^+}_2$ where $\sigma_i, i=1,\ldots,r=\mbox{rank}(\W)$, are the singular values of $\W$ \cite{Beltran1991}.}
 has been used as the objective function for optimization of the trajectories.  The 
independent coordinates of the robot, $z_1,\ldots,z_6$, have been parametrized using a finite Fourier series, as proposed in \cite{Swevers1996}.
 A total of $100$ sampling data points evenly covering a full period are extracted from each optimized trajectory. In order to obtain a characteristic sample set
 that sufficiently excites the system under the defined optimization conditions, the sampled data  from $10$ different optimized trajectories is used for the estimation.
 The MATLAB Optimization Toolbox function \verb!fmincon! has been used to perform the actual optimization.
The rotation angles and the angular velocity ranges of the actuators have been limited using linear inequality constraints.
For reference, the details of the trajectory optimization setup are summarized in Table~\ref{tab::trajectory_characteristics_puma}.
To give a graphical idea, one of the optimized trajectories is shown in Fig.~\ref{fig:puma_trajectory_coordinates}.

\begin{table}[!ht]
\caption{HME: Exciting Trajectory definition}
\label{tab::trajectory_characteristics_puma}
\centering
\begin{minipage}[b]{0.4\linewidth}\centering
\footnotesize
\begin{tabular}{lc}
\hline
Optimization criterion			&	Cond. Number				\\
% \hline
\# actuated joints			&	$6$						\\
% \hline
\# harmonics				&	$4$						\\
% \hline
\# trajectory parameters		&	$54$						\\
% \hline
\# sample points per traj.			&	$100$						\\
% \hline
\# linear inequality constraints	&	$2400$						\\
% \hline
\# non-linear inequality constraints	&	$0$						\\
%  \hline
% \end{tabular} 
% \end{minipage}
% \begin{minipage}[b]{0.4\linewidth}\centering
% \footnotesize
% \begin{tabular}{lc}
%  \hline
main trajectory period  	&	$2\pi$~s					\\
% \hline
$z_{min}$				&	$-\pi/2$~rad					\\
% \hline
$z_{max}$				&	$\phantom{-}\pi/2$~rad			\\
% \hline
$\dot{z}_{min}$				&	$-1.45$~rad/s				\\
% \hline
$\dot{z}_{max}$				&	$\phantom{-}1.45$~rad/s			\\
% \hline
\# Estimation Trajectories 	&	$10$						\\
% \hline
\# Validation Trajectories  &	$1$						\\
\hline
\end{tabular} 
\end{minipage}
\end{table}
%

%--------------------------------------------------
\subsubsection*{Performance of the parameter-reduced models} \label{sec::reduced_mode_efficiency_results_puma}

\begin{figure}[!ht]
  \begin{center}
    \includegraphics[width=10.32cm]{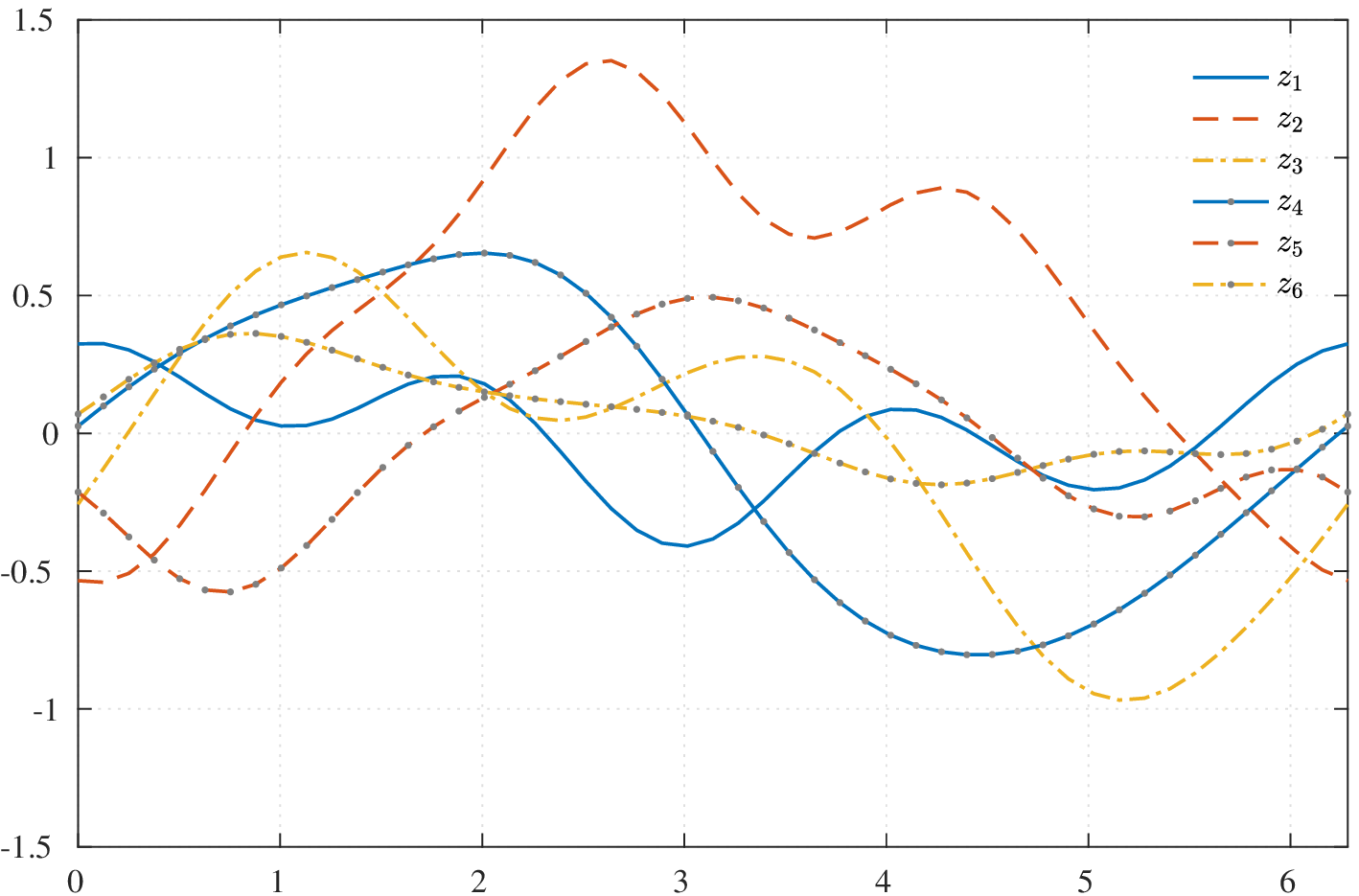}
    \vspace*{-3mm}
    \caption{HME: An example of a characteristic trajectory ($\cal V$).  \textit{Time} ($\mbox{s}$) is shown on the x-axis.} 
    \label{fig:puma_trajectory_coordinates} 
  \end{center}
% \end{figure}
% %
% \begin{figure}[!ht]
  \begin{center}
    \includegraphics[width=10.32cm]{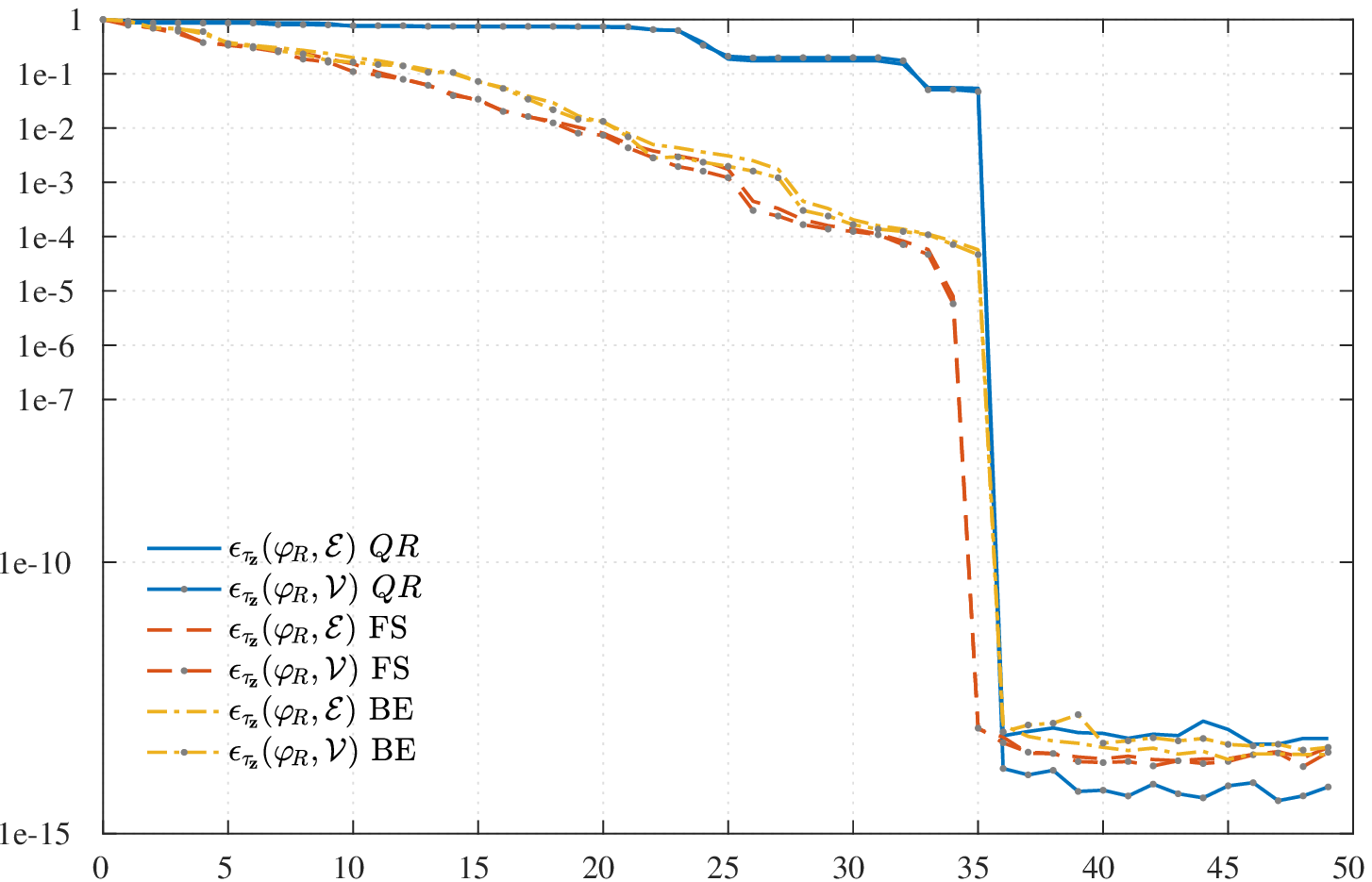}
    \vspace*{-3mm}
    \caption{HME: $\epsilon_{\vtau_{\!\!\vz}}(\vphi_{\!R},{\cal E})$ and $\epsilon_{\vtau_{\!\!\vz}}(\vphi_{\!R},{\cal V})$ vs. $n_{\vphi_{\!R}}$ for the $QR$, FS and BE heuristics.} 
    \label{fig:puma560_epsilon_tau_n_phi} 
  \end{center}
% \end{figure}
% %
% \begin{figure}[!ht]
  \begin{center}
    \includegraphics[width=10.32cm]{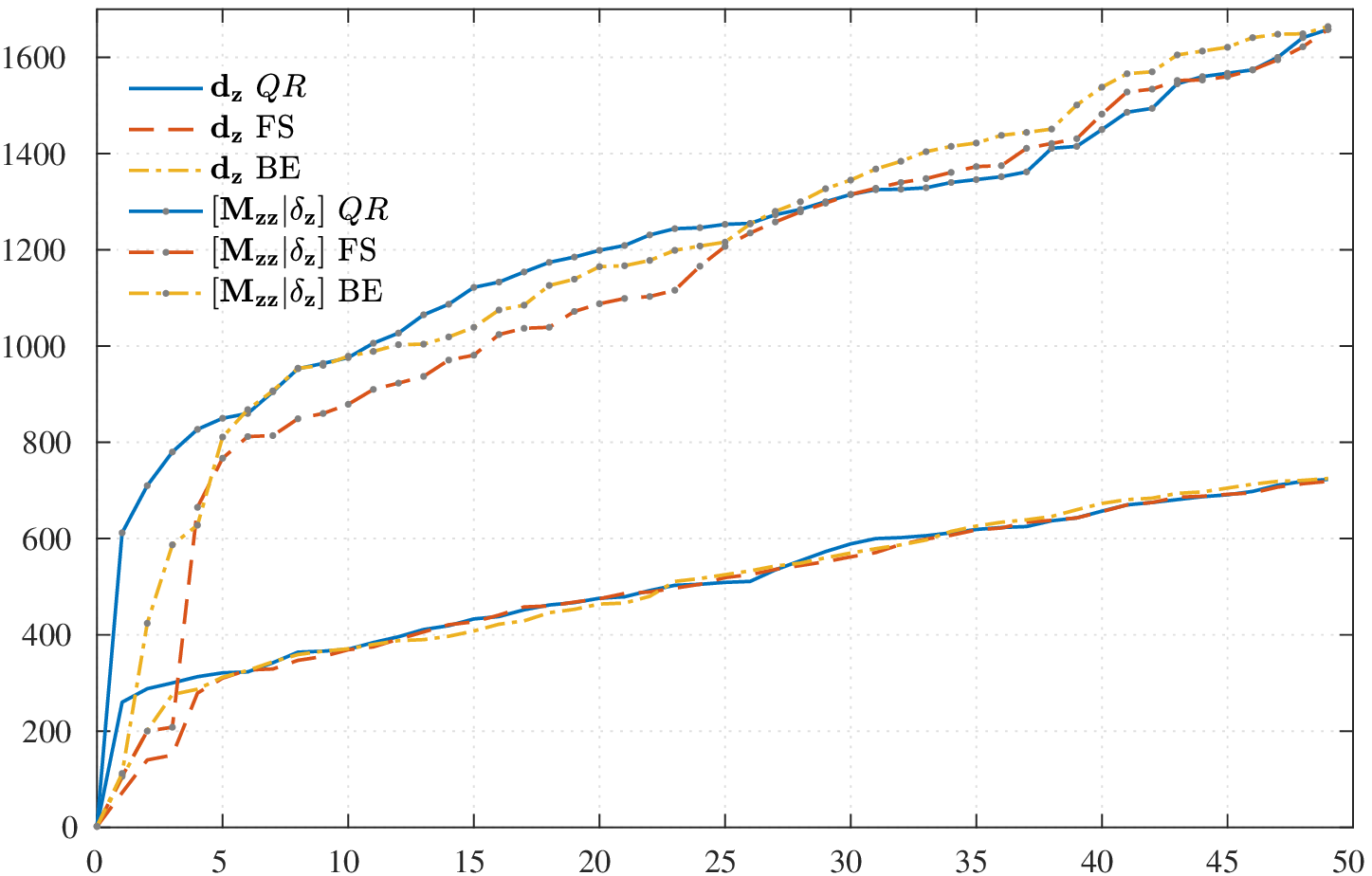}
    \vspace*{-3mm}
    \caption{HME: $n_{op}$ vs. $n_{\vphi_{\!R}}$ for functions $\mathbf{d}_{\vz}$ and $[\M_{\!\vz\vz}|\vdelta_{\!\vz}]$  for the $QR$, FS and BE heuristics.} 
    \label{fig:puma560_n_op_n_phi} 
  \end{center}
\end{figure}

\begin{figure}[!ht]
  \begin{center}
    \includegraphics[width=10.32cm]{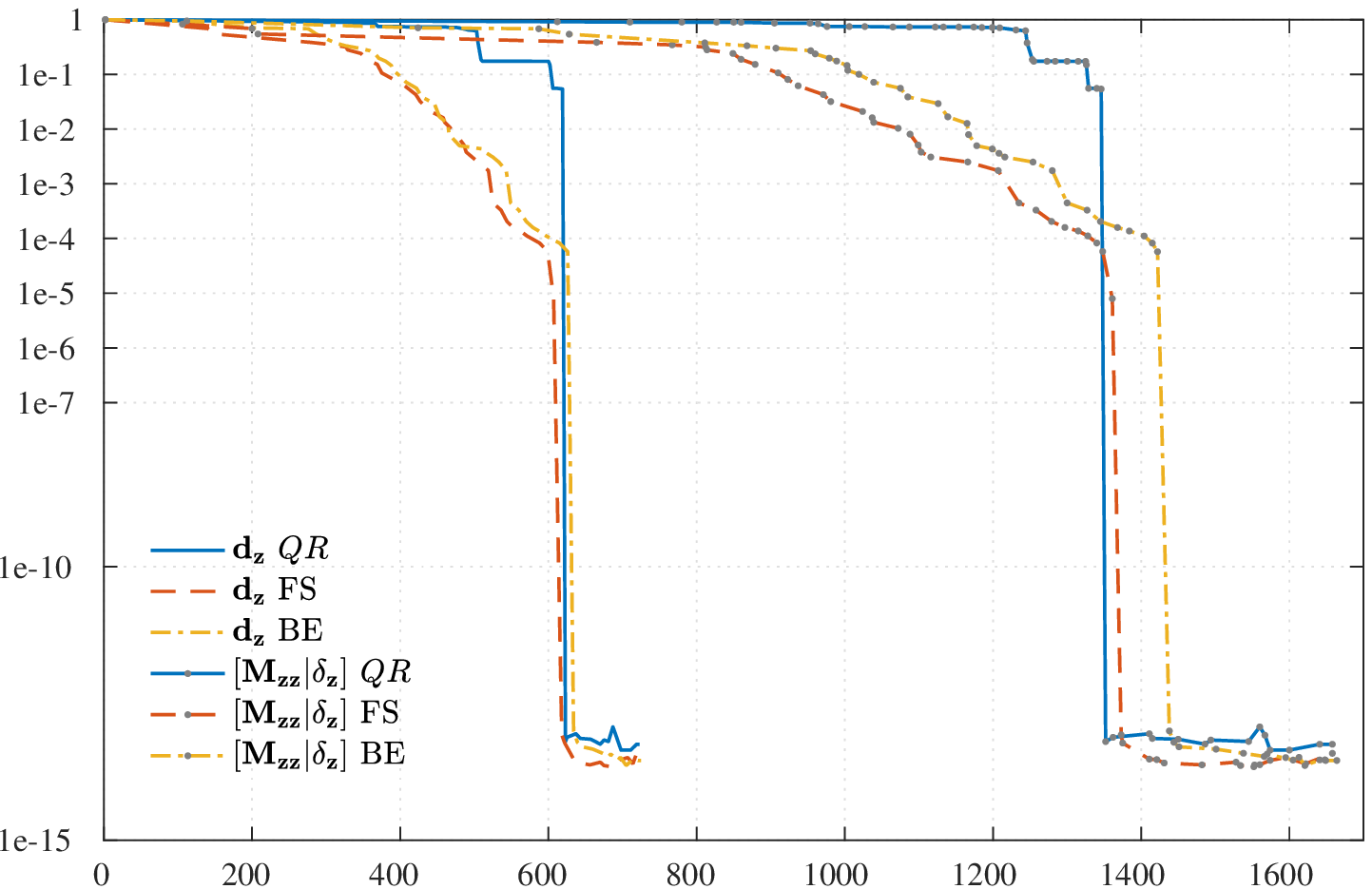}
    \vspace*{-3mm}
    \caption{HME: $\epsilon_{\vtau_{\!\!\vz}}(\vphi_{\!R},{\cal E})$ vs. $n_{op}$ for functions $\mathbf{d}_{\vz}$ and $[\M_{\!\vz\vz}|\vdelta_{\!\vz}]$ for the $QR$, FS and BE heuristics.} 
    \label{fig:puma560_epsilon_tau_n_op} 
  \end{center}
% \end{figure}
% %
% \begin{figure}[!ht]
  \begin{center}
    \includegraphics[width=10.32cm]{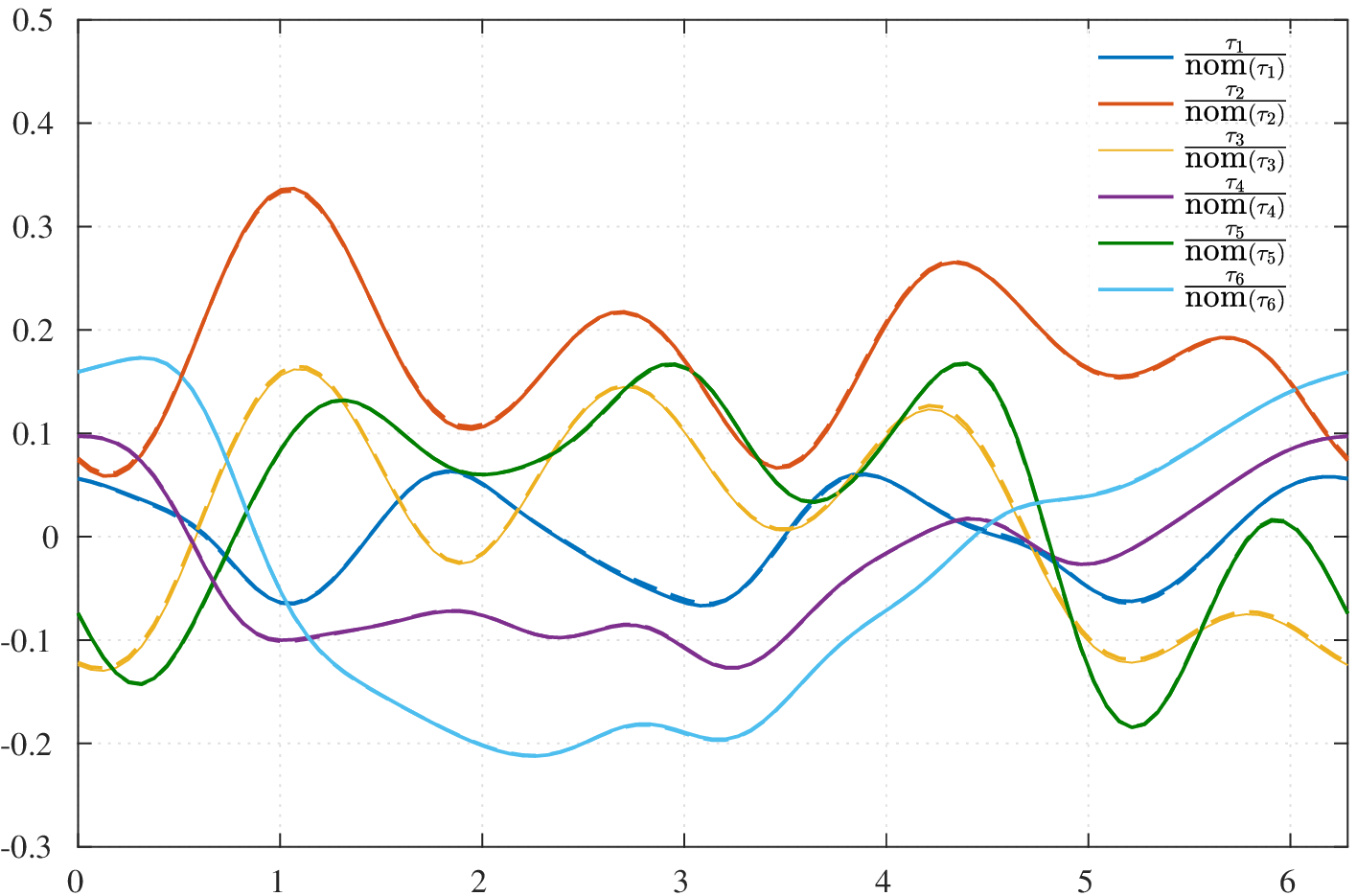}
    \vspace*{-3mm}
    \caption{HME: Selected model (FS $n_{\vphi_{\!R}}=18$) $\frac{{\tau_i(\vphi_{\!R})}}{\mbox{nom}({\tau_i})}$ ($--$) and $\frac{{\tau_i(\vphi)}}{\mbox{nom}({\tau_i})}$ ($-$) vs. $t$ ($\mbox{s}$).}
    \label{fig:hexaglide_tau_norm_validation} 
    \label{fig:puma560_tau_norm_validation} 
  \end{center}
% \end{figure}
% %
% \begin{figure}[!ht]
  \begin{center}
    \includegraphics[width=10.32cm]{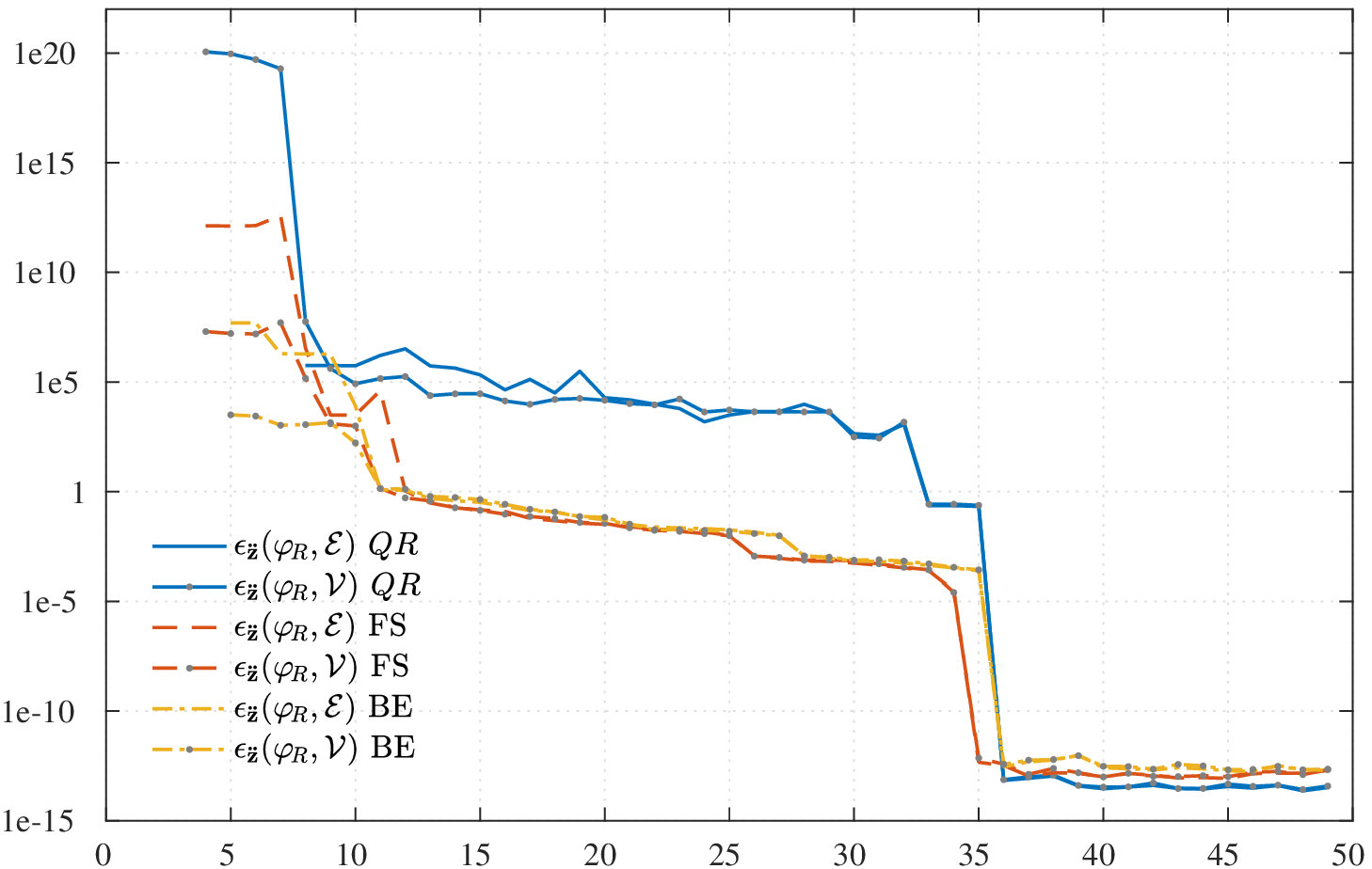}
    \vspace*{-3mm}
    \caption{HME: $\epsilon_{\vddz}(\vphi_{\!R},{\cal E})$ and $\epsilon_{\vddz}(\vphi_{\!R},{\cal V})$  vs. $n_{\vphi_{\!R}}$ for the $QR$, FS and BE heuristics.} 
    \label{fig:puma560_epsilon_ddq_n_phi} 
  \end{center}
\end{figure}

The performance of the previously proposed  algorithms is shown in Figs.~\ref{fig:puma560_epsilon_tau_n_phi},  \ref{fig:puma560_n_op_n_phi}, \ref{fig:puma560_epsilon_tau_n_op},  \ref{fig:puma560_tau_norm_validation} and \ref{fig:puma560_epsilon_ddq_n_phi}.
Fig.~\ref{fig:puma560_epsilon_tau_n_phi} represents the normalized prediction error for the parameter-reduced inverse dynamic models $\epsilon_{\vtau_{\!\!\vz}}$, in terms
of the number of parameters $n_{\vphi_{\!R}}$, using the three proposed heuristics for both the estimation, $\cal E$, and validation, $\cal V$, data sets.
Fig.~\ref{fig:puma560_n_op_n_phi} shows the number of operations, $n_{op}$, required to compute the functions $\mathbf{d}_{\vz}(\vz,\vdz,\vddz,\vphi_R)$ and $[\M_{\!\vz\vz}|\vdelta_{\!\vz}](\vz,\vdz,\vddz,\vphi_{\!R})$, in terms
of the number of parameters $n_{\vphi_{\!R}}$, for the three proposed heuristics.
The number of operations, $n_{op}$, refers to the total
number of \textit{non-repeated} sums, products and function evaluations required.
Fig.~\ref{fig:puma560_epsilon_tau_n_op} represents the normalized prediction error $\epsilon_{\vtau_{\!\!\vz}}$ in terms of the number of operations, $n_{op}$ for the same functions and heuristics as in Fig.~\ref{fig:puma560_n_op_n_phi}. 
Fig.~\ref{fig:puma560_tau_norm_validation} compares the normalized torque for the full model and the parameter-reduced model with $n_{\vphi_R}\!=18$ obtained with the \textit{Forward Selection} heuristic for the validation trajectory.
Fig.~\ref{fig:puma560_epsilon_ddq_n_phi}  shows the normalized error for the parameter-reduced DDM, $\epsilon_{\vddz}$, in terms
of the number of parameters $n_{\vphi_{\!R}}$, using the proposed heuristics for
the estimation, $\cal E$, and validation, $\cal V$, data sets.

Finally, for completeness, the parameter ordering corresponding to the different heuristics analyzed is presented in Table~\ref{tab::puma560_parameter_ordering}. For a given heuristics row, the set of parameters
corresponding to the parameter-reduced model with $n_{\vphi_{\!R}}$ parameters are those related to the column indexes $1,\ldots,n_{\vphi_{\!R}}$.
\renewcommand{\arraystretch}{1.7}
\begin{table}[!ht]
\caption{HME: Parameter ordering for the $QR$, FS and BE heuristics.}
\label{tab::puma560_parameter_ordering}
\centering
% \begin{minipage}[b]{0.5\linewidth}\centering
\footnotesize
\begin{tabular}{ccccccccccc}
\hline
$Par. \#$ & 1 & 2 & 3 & 4 & 5 & 6 & 7 & 8 & 9 & 10  \\
\hline
BE & $d_y^3$ & $m^6$ & $I_{yy}^6$ & $I_{xx}^6$ & $d_y^6$ & $d_x^6$ & $d_y^4$ & $d_y^5$ & $I_{zz}^2$ & $I_{zz}^6$ \\
% \hline
FS & $m^3$ & $I_{zz}^5$ & $I_{zz}^3$ & $d_y^6$ & $d_x^6$ & $d_y^5$ & $d_x^2$ & $d_y^4$ & $I_{zz}^4$ & $d_y^3$ \\
% \hline
QR & $d_y^6$ & $d_x^6$ & $I_{xy}^6$ & $I_{yz}^6$ & $I_{xz}^6$ & $I_{zz}^6$ & $d_y^5$ & $d_x^5$ & $I_{yy}^6$ & $I_{xx}^6$ \\
 \hline
 \vspace{-0.15cm}\\
\hline
$Par. \#$ & 11 & 12 & 13 & 14 & 15 & 16 & 17 & 18 & 19 & 20  \\
\hline
BE & $I_{zz}^4$ & $I_{xx}^5$ & $I_{zz}^1$ & $m^3$ & $d_x^3$ & $d_x^4$ & $I_{xx}^2$ & $d_x^5$ & $I_{yy}^4$ & $d_z^3$ \\
% \hline
FS & $I_{zz}^6$ & $I_{xx}^3$ & $I_{xz}^2$ & $d_x^4$ & $I_{xx}^4$ & $d_x^5$ & $I_{yy}^3$ & $d_y^2$ & $m^4$ & $I_{yy}^5$ \\
% \hline
QR & $I_{yz}^5$ & $I_{xz}^5$ & $d_y^4$ & $I_{xy}^5$ & $d_x^4$ & $I_{xx}^5$ & $I_{yz}^4$ & $I_{xz}^4$ & $I_{yy}^5$ & $I_{xy}^4$ \\
 \hline
 \vspace{-0.15cm}\\
\hline
$Par. \#$ & 21 & 22 & 23 & 24 & 25 & 26 & 27 & 28 & 29 & 30  \\
\hline
BE & $d_y^2$ & $I_{yz}^2$ & $I_{xy}^3$ & $I_{xy}^2$ & $I_{zz}^3$ & $I_{yz}^6$ & $I_{xy}^6$ & $I_{xz}^6$ & $I_{xz}^5$ & $I_{yz}^5$ \\
% \hline
FS & $I_{yz}^2$ & $I_{yy}^2$ & $I_{xy}^2$ & $I_{yz}^6$ & $I_{xy}^6$ & $I_{xz}^6$ & $I_{xz}^5$ & $I_{yz}^5$ & $I_{xy}^5$ & $I_{yz}^4$ \\
% \hline
QR & $I_{yy}^4$ & $d_y^3$ & $d_x^3$ & $d_x^2$ & $I_{zz}^4$ & $d_y^2$ & $I_{xz}^3$ & $I_{yz}^3$ & $I_{xy}^3$ & $I_{xz}^2$ \\
\hline
% \end{tabular}
% \end{minipage}
% \begin{minipage}[b]{0.5\linewidth}\centering
% \footnotesize
% \begin{tabular}{|c|cccccccccc|}
\vspace{-0.15cm}\\
\hline
$Par. \#$ & 31 & 32 & 33 & 34 & 35 & 36 & 37 & 38 & 39 & 40  \\
\hline
BE & $I_{xy}^5$ & $I_{yz}^4$ & $I_{xz}^4$ & $I_{xz}^3$ & $I_{yz}^3$ & $I_{xy}^4$ & $I_{yy}^2$ & $I_{xz}^2$ & $d_z^6$ & $d_z^4$ \\
% \hline
FS & $I_{xz}^4$ & $I_{xz}^3$ & $I_{yz}^3$ & $I_{xy}^4$ & $I_{xy}^3$ & $I_{zz}^2$ & $d_z^5$ & $I_{yy}^4$ & $I_{xx}^5$ & $d_z^6$ \\
% \hline
QR & $I_{yz}^2$ & $I_{zz}^1$ & $I_{zz}^2$ & $I_{xy}^2$ & $I_{xx}^3$ & $I_{xx}^2$ & $I_{zz}^5$ & $d_z^6$ & $I_{zz}^3$ & $d_z^5$ \\
 \hline
\vspace{-0.15cm}\\
\hline
$Par. \#$ & 41 & 42 & 43 & 44 & 45 & 46 & 47 & 48 & 49 & \phantom{mm}  \\
\hline
BE & $m^4$ & $I_{xx}^4$ & $d_z^5$ & $I_{zz}^5$ & $I_{xx}^3$ & $m^5$ & $I_{yy}^3$ & $d_x^2$ & $I_{yy}^5$ & $d_z^2$ \\	%& \phantom{mm} \\
% \hline
FS & $d_z^4$ & $I_{xx}^2$ & $d_x^3$ & $I_{zz}^1$ & $I_{yy}^6$ & $I_{xx}^6$ & $d_z^3$ & $m^5$ & $m^6$ & $d_z^2$ \\	%& \phantom{mm} \\
% \hline
QR & $d_z^4$ & $I_{xx}^4$ & $m^4$ & $m^3$ & $I_{yy}^2$ & $I_{yy}^3$ & $d_z^3$ & $m^6$ & $m^5$ & $d_z^2$ \\		%& \phantom{mm} \\
 \hline
 \vspace{-0.15cm}\\
\hline
$Par. \#$ & 51 & 52 & 53 & 54 & 55 & 56 & 57 & 58 & 59 & 60 \\
\hline
BE    & $m^2$      & $I_{yz}^1$ & $I_{xz}^1$ & $I_{xy}^1$ & $I_{yy}^1$ & $I_{xx}^1$ & $d_z^1$    & $d_x^1$    & $d_x^1$    & $m^1$ \\
% \hline
FS    & $m^2$      & $I_{yz}^1$ & $I_{xz}^1$ & $I_{xy}^1$ & $I_{yy}^1$ & $I_{xx}^1$ & $d_z^1$    & $d_x^1$    & $d_x^1$    & $m^1$ \\
% \hline
$QR$ & $m^2$      & $I_{yz}^1$ & $I_{xz}^1$ & $I_{xy}^1$ & $I_{yy}^1$ & $I_{xx}^1$ & $d_z^1$    & $d_x^1$    & $d_x^1$    & $m^1$ \\
\hline
\end{tabular}

% \end{minipage}
\end{table}
\renewcommand{\arraystretch}{1}

\afterpage{\clearpage}

\pagebreak
%
%--------------------------------------------------
\subsection{Low-mobility multibody example} \label{sec::hexaglide_robot}

\begin{figure}[!ht]
  \begin{center}
    \includegraphics[width=10.32cm]{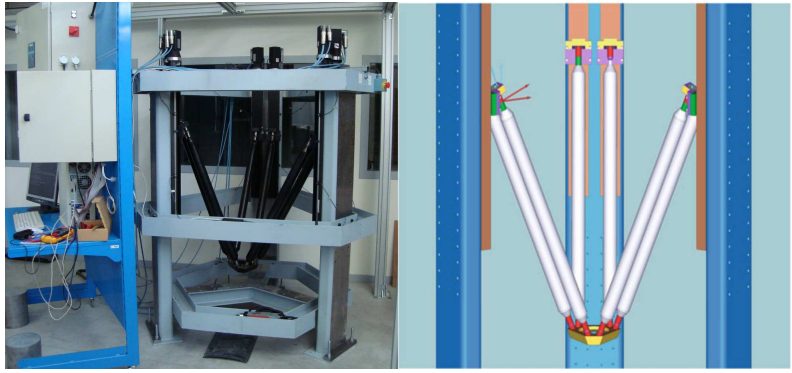}   
    \caption{Hexaglide: left) Actual prototype, right) CAD model.} 
    \label{fig:Hexaglide_foto_3d} 
  \end{center}
\end{figure}

The particular difficulties of parallel mechanism identification  are well known. For example, in \cite{Valero2013} the problems
associated with an automobile suspension system are analyzed. These are frequently related to the presence of closed loops that
 limit  the mobility of some bodies to a significant extent. In this study, the $6$-PUS Hexaglide \cite{Ros2013}, shown in Fig.~\ref{fig:Hexaglide_foto_3d}, is
the selected low-mobility multibody system example (LME).

A total of $24$ generalized coordinates, $\vq$, are used to model the system: the heights of the $6$ guides, $\vz=[z_1,\ldots,z_6]$, that are the independent/actuated coordinates, $2$ relative rotations
of the universal joints for each of the $6$ arms, and $6$ absolute coordinates for the head (the Cartesian coordinates of the head center, and its Euler angles).
For reference purposes, the  kinematic parameters  of the prototype are shown in Table~\ref{tab::hexaglide_kin_param}.
The closed-loop nature of the system requires the enforcement of $18$ constraints ($3$ for each $S$ bar-head joint), leading to a total of $6$ \textit{d.o.f}.
As before, the \textit{Principle of Virtual Power} has been used to obtain the dynamic equations. The inertial parameters of the bars, referenced to the U joint with the
carriages,  and those of the Head, referenced to the center of the S joints, are given in
Table~\ref{tab::hexaglide_dyn_param}.
The required constraint enforcement is performed using Lagrange multipliers $\boldsymbol{\lambda}$, leading to the set of dynamic equations
\begin{equation}
 \mathbf{d}_{\!\vq}=\M_{\!\vq\vq} \vddq  - \vdelta_{\!\vq} + \boldsymbol{\phi}_{\mathbf{q}}^T \boldsymbol{\lambda}    = \vtau_{\!\!\vq},
 \label{eq::M_prime}
\end{equation}
where $\boldsymbol{\phi}_{\vq}=\frac{\partial \boldsymbol{\phi}}{\partial \vq}$ is the constraint Jacobian. This equation can be rearranged as
\begin{equation}
 \mathbf{d}_{\!\vq}=\K_{\!\vq\vphi} \vphi + \boldsymbol{\phi}_{\vq}^T \boldsymbol{\lambda} = \vtau_{\!\!\vq}.
 \label{eq::K_prime}
\end{equation}

The generalized coordinate vector is ordered as $\mathbf{q}=\left[\mathbf{d}^T,\vz^T\right]^T$, where $\vz$ is the independent
coordinate set  (linear actuator coordinates). Accordingly, the Jacobian columns are ordered as 
$\boldsymbol{\phi}_{\vq}=\left[\boldsymbol{\phi}_\mathbf{d},\boldsymbol{\phi}_{\vz}\right]$.

Now, noting that 
\begin{equation}
 \mathbf{R}=
 \begin{bmatrix}
  -\boldsymbol{\phi}_{\mathbf{d}}^{-1}\boldsymbol{\phi}_{\vz} \\
  \mathbf{1}
 \end{bmatrix}
\end{equation}
is an orthogonal complement of $\boldsymbol{\phi}_{\vq}$,
 $\boldsymbol{\phi}_{\vq}\mathbf{R}=\mathbf{0}$,
left-multiplying Eqs.~(\ref{eq::M_prime}) and (\ref{eq::K_prime}) by $\mathbf{R}^T$ and substituting
\begin{equation}
 \ddot{\mathbf{q}}=\mathbf{R} \vddz + \begin{bmatrix}
                                       \boldsymbol{\phi}_{\mathbf{d}}^{-1} \boldsymbol{\gamma}\\
                                       \mathbf{0}
                                      \end{bmatrix},
\end{equation}
where $\boldsymbol{\gamma}=\boldsymbol{\phi}_{\vq} \vddq-\ddot{\boldsymbol{\phi}}$, in Eq.~(\ref{eq::M_prime}),  these equations take the form of
 Eqs.~(\ref{eq::lagrange_dynamic_model}) and (\ref{eq::lp_dynamic_model}). This leads to the formulation on independent coordinates of the dynamic
  functions
 \begin{eqnarray}
 \mathbf{d}_{\vz} &=& \mathbf{R}^{T} \mathbf{d}_{\vq}, \\
 \M_{\!\vz\vz} &=& \mathbf{R}^{T}\M_{\!\vq\vq}\mathbf{R}, \\
 \vdelta_{\vz} &=&\mathbf{R}^{T} \left( \vdelta_{\vq} - \begin{bmatrix}
                                       \boldsymbol{\phi}_{\mathbf{d}}^{-1} \boldsymbol{\gamma}\\
                                       \mathbf{0}
                                      \end{bmatrix} \right), \\
 \K_{\!\vz\vphi} &=& \mathbf{R}^{T} \K_{\!\vq\vphi}, \mbox{and}\\
 \vtau_{\!\!\vz} &=& \mathbf{R}^{T} \vtau_{\!\!\vq},
\end{eqnarray}
chosen to describe the algorithms introduced in this paper.
The state, $(\vq,\vdq)$, is determined as a function of the independent coordinates and velocities,
$(\vz,\dot{\vz})$, making use of the constraint equations at the level of position and velocity.
\begin{table}[!ht]
\caption{Kinematic parameters for Hexaglide.}
\label{tab::hexaglide_kin_param}
\footnotesize
\begin{center}
\begin{tabular}{lcc}
\hline
Bar length & $L$ & $1.0000$~m\\
% \hline
Linear guide separation & $e$ & $0.1365$~m\\
% \hline
Head $S$ joint separation & $e$ & $0.1365$~m\\
% \hline
Frame symmetry  axis to $U$ joint & $R$ & $0.4840$~m\\
% \hline
Head symmetry axis to $S$ joint & $r$ & $0.0730$~m\\
% \hline
Equilateral symmetry & $\alpha$ & $2 \pi/3$~rad\\
\hline
\end{tabular}
\end{center}
% \end{table}
% %
% \begin{table}[!ht]
\caption{Dynamic parameters for Hexaglide.}
\label{tab::hexaglide_dyn_param}
\begin{center}
\footnotesize
\begin{tabular}{lcccccccccc}
\hline
{Body} & $m$ & $d_x$ & $d_y$ & $d_z$ & $I_{xx}$ & $I_{xy}$ & $I_{yy}$ & $I_{xz}$ & $I_{zz}$ & $I_{yz}$ \\
% \hline
{Unit} & kg & kg~m & kg~m & kg~m & kg~m$^2$ & kg~m$^2$ & kg~m$^2$ & kg~m$^2$ & kg~m$^2$ & kg~m$^2$ \\
\hline
% \hline
Head &  $6.697$ &  $0.07$ &  $0.07$ &  $-0.238$ &  $0.0283$ &  $0.001$ &  $0.028$ &  $0.000$  &  $0.038$  &  $0.000$ \\ 
% \hline
Bar &  $5.804$ &  $0.03$  &  $0.03$	 &  $-1.469$  &  $1.044$  &  $0.000$  &  $1.044$  &  $0.014$    &  $0.002$  &  $0.014$ \\
\hline
\end{tabular}
\end{center}
\end{table}
As with the previous example, the symbolic library in \cite{Ros2007} has been used to obtain the equations.

As the parameter reduction procedure does not affect the kinematic model
 and, therefore, the computation of $\mathbf{R}$, we propose to show the computational
 complexity of the parameter-reduced models of this example in terms of the number of operations required
 to compute $\mathbf{d}_{\vq}(\!\vq,\vdq,\vddq,\vphi_{\!R})$ and $[\M_{\!\vq\vq}|\vdelta_{\!\vq}](\!\vq,\vdq,\vddq,\vphi_{\!R})$. 
 These are the matrices directly exported by the  symbolic library. 
 
The nominal forces of the actuators, $\mbox{nom}({\vtau}_{\!\!\vz})=1.7~\frac{2~\pi}{10^{-2}}\left[1,1,1,1,1,1\right]~\mbox{N}$, are used to define the normalized prediction error, $\epsilon_{\vtau}$.

 \subsubsection*{Estimation trajectories}

The trajectories are obtained using the same procedure followed for the previous example. The $6$ independent coordinates, $z_1,\ldots,z_6$,
have been parametrized using finite Fourier series.
 In this case, the characteristic sample set uses data from $25$ different optimized trajectories. For each trajectory $400$ sample points are extracted.
% To perform the actual optimization the \verb!fmincon! function in MATLAB optimization toolbox has been used.
% The rotation angle and the angular velocity ranges of the actuators have been limited using linear inequality constraints.
% To avoid collision with the support structure and to avoid singular configurations, non-linear inequality constraints are used: the center of the platform
% distance to the symmetry axis of the robot should be $\le (R-r) 0.9$, and the angle between the platform the bars should be $\ge 5^{o}$.
For reference purposes, the details of the trajectory optimization setup are summarized in Table~\ref{tab::trajectory_characteristics_hexaglide}.
\begin{table}[!ht]
\caption{LME: Exciting Trajectory definition}
\label{tab::trajectory_characteristics_hexaglide}
\centering
\begin{minipage}[b]{0.4\linewidth}\centering
\footnotesize
\begin{tabular}{lcc}
\hline
Optimization criterion			&		Cond. Number	\\
% \hline
\# actuated joints			&		$6$		\\
% \hline
\# harmonics				&		$2$		\\
% \hline
\# trajectory parameters		&		$30$		\\
% \hline
\# sample points per traj.			&		$400$		\\
% \hline
\# linear inequality constraints	&		$2400$		\\
% \hline
\# non-linear inequality constraints	&		$700$		\\
 \hline
\end{tabular} 
\end{minipage}
\begin{minipage}[b]{0.4\linewidth}\centering
\footnotesize
\begin{tabular}{lc}
\hline
main trajectory period  	&		$2\pi$~s		\\
% \hline
$z_{min}$				&		$1$~m 		\\
% \hline
$z_{max}$				&		$2$~m 		\\
% \hline
$\dot{z}_{min}$				&		$-1.00$~m/s	\\
% \hline
$\dot{z}_{max}$				&	 $\phantom{-}1.00$~m/s	\\
% \hline
\# Estimation Trajectories 	&		$25$		\\
% \hline
\# Validation Trajectories 	&		$1$		\\
\hline
\end{tabular} 
\end{minipage}
\end{table}
To give a graphical idea, one of the optimized trajectories is shown in Fig.~\ref{fig:hexaglide_trajectory_coordinates}.

%
% Trajectories have been optimized to minimize the condition number of matrix $\W_B$ in Eq.~(\ref{eq::perfect_model_reduction_W}). 
% As the procedures eliminate or add one parameter at a time, all the intermediate models have been saved so that $p$ different reduced models are obtained, each of them with a different number of parameters. 
% For each of them, the \red{bias in prediction} has been calculated in order to observe how it increases as the models depend on less parameters.
% 
% The specific characteristics of the excitation trajectories of both systems are shown in Table (\ref{tab::trajectory_characteristics}). The coordinates of two of the optimized trajectories are shown in Figs. (\ref{fig:puma_trajectory_coordinates},\ref{fig:hexaglide_trajectory_coordinates}).
% 
% The results obtained applying these algorithms to the Puma and Hexaglide robots are the ones in Figs. (\ref{fig:puma560_epsilon_tau_n_phi}) and (\ref{fig:hexaglide_epsilon_tau_n_phi}), respectively. The solid lines represent the error of the reduced model calculating the forces of the reduction trajectories. The dashed lines represent the errors of the reduced models (obtained using the reduction trajectories) calculating the forces of the validation trajectories. The three different colors (red, green and blue) denote the three different reduction methods presented in this paper (\textit{Forward Selection}, forward selection and QR decomposition).

%--------------------------------------------------
\subsubsection*{Performance of the  parameter-reduced models}
\begin{figure}[!ht]
  \begin{center}
    \includegraphics[width=10.32cm]{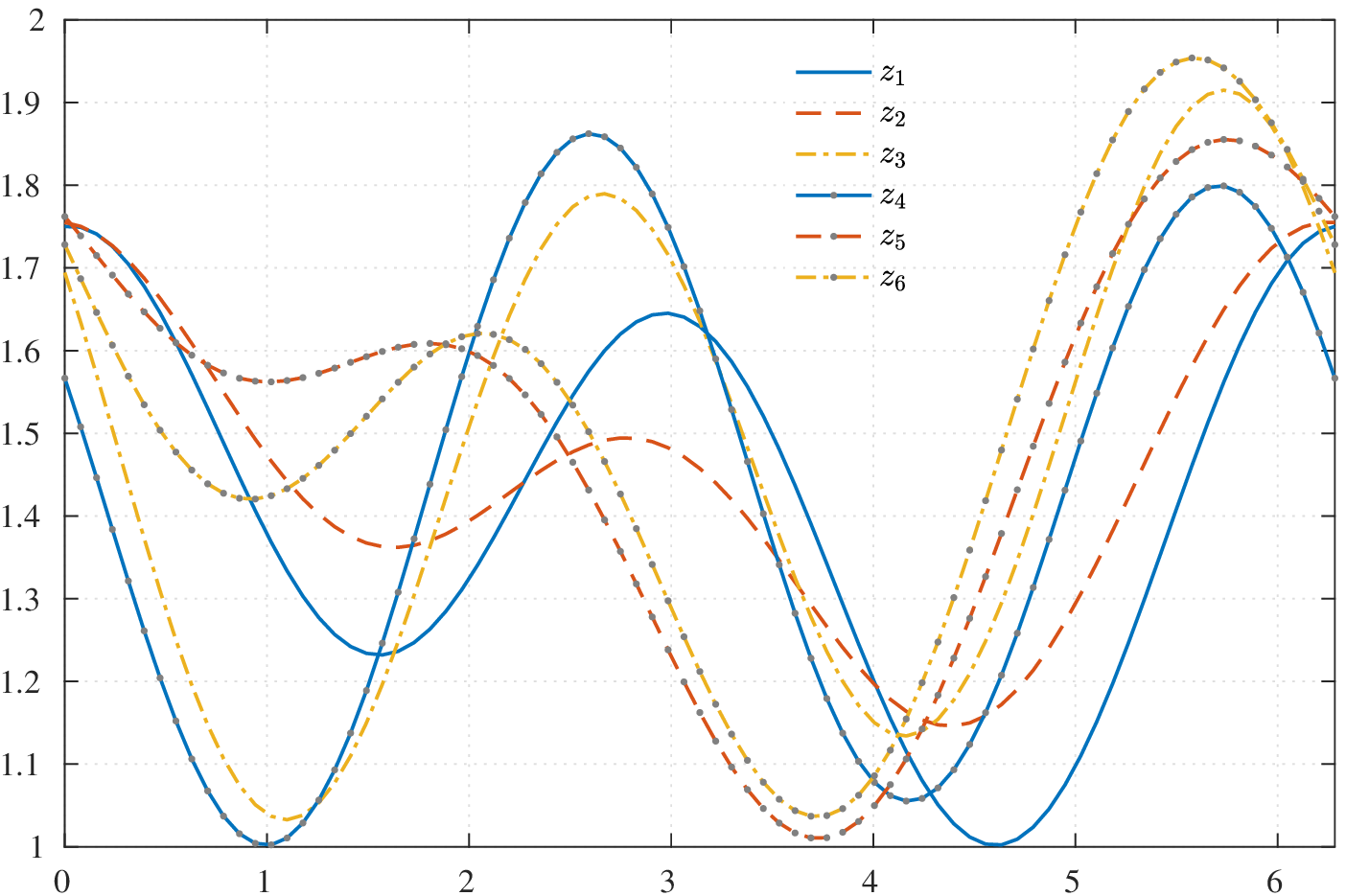} 
    \vspace*{-3mm}
    \caption{LME: An example of characteristic trajectory ($\cal V$). \textit{Time} ($\mbox{s}$) is shown on the x-axis.}
    \label{fig:hexaglide_trajectory_coordinates} 
  \end{center}
% \end{figure}
% %
% \begin{figure}[!ht]
  \begin{center}
    \includegraphics[width=10.32cm]{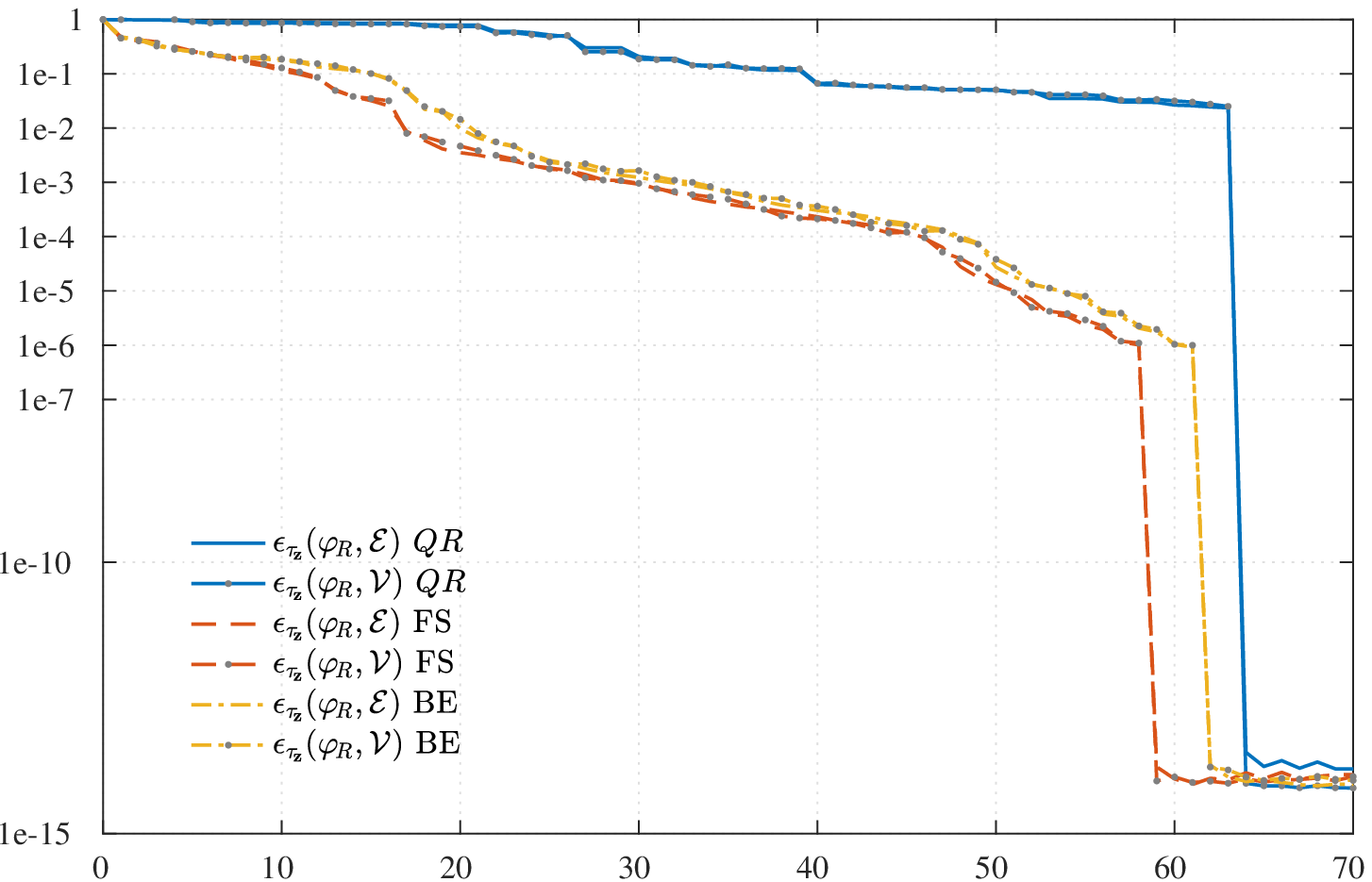}
    \vspace*{-3mm}
    \caption{LME: $\varepsilon_{\vtau_{\!\!\vz}}(\vphi_{\!R},{\cal E})$ and $\varepsilon_{\vtau_{\!\!\vz}}(\vphi_{\!R},{\cal E})$ vs. $n_{\vphi_{\!R}}$ for the  $QR$ , FS, and BE  heuristics.}  
    \label{fig:hexaglide_epsilon_tau_n_phi} 
  \end{center}
% \end{figure}
% %
% \begin{figure}[!ht]
  \begin{center}
    \includegraphics[width=10.32cm]{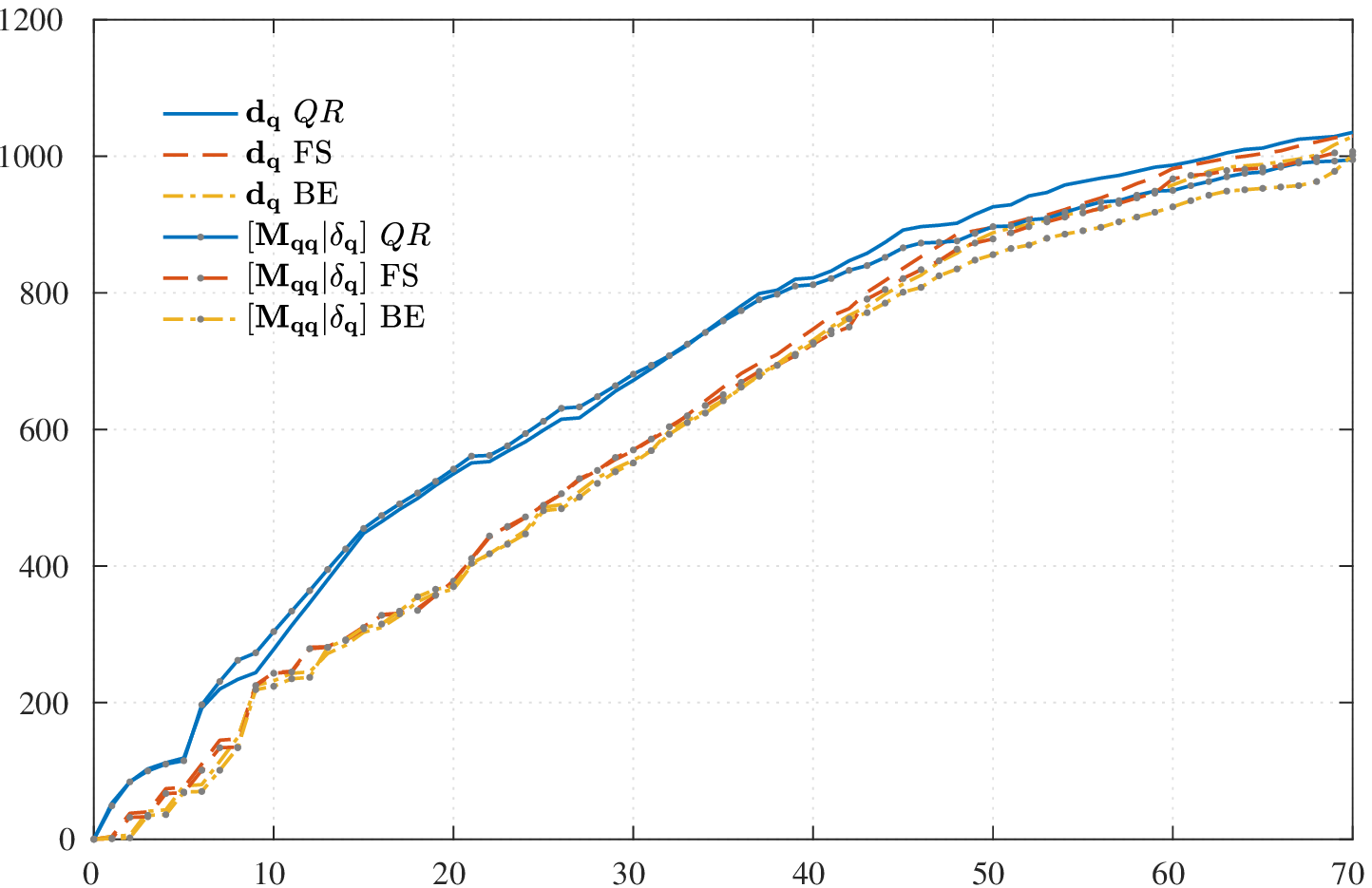} 
    \vspace*{-3mm}
    \caption{LME: $n_{op}$ vs. $n_{\vphi_{\!R}}$ for functions $\mathbf{d}_{\!\vq}$ and $[\M_{\!\vq\vq}|\vdelta_{\!\vq}]$  for the $QR$, FS and BE heuristics.} 
    \label{fig:hexaglide_n_op_n_phi} 
  \end{center}
\end{figure}

\begin{figure}[!ht]
  \begin{center}
    \includegraphics[width=10.32cm]{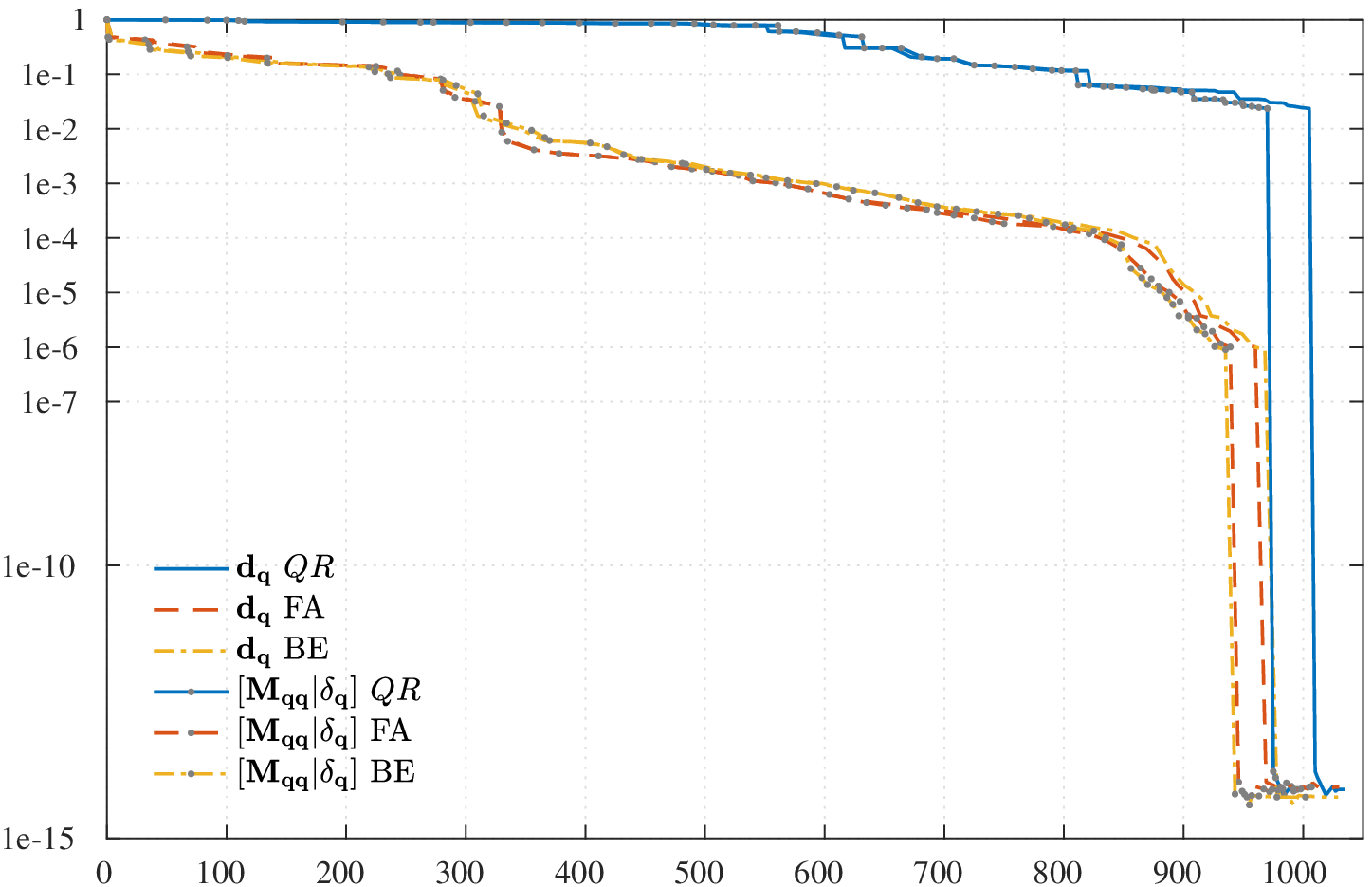} 
    \vspace*{-3mm}
    \caption{LME: $\epsilon_{\vtau_{\!\!\vz}}(\vphi_{\!R},{\cal E})$ vs. $n_{op}$ for functions $\mathbf{d}_{\!\vq}$ and $[\M_{\!\vq\vq}|\vdelta_{\!\vq}]$ for the $QR$, FS and BE heuristics.} 
    \label{fig:hexaglide_epsilon_tau_n_op} 
  \end{center}
% \end{figure}
% %
% \begin{figure}[!ht]
  \begin{center}
    \includegraphics[width=10.32cm]{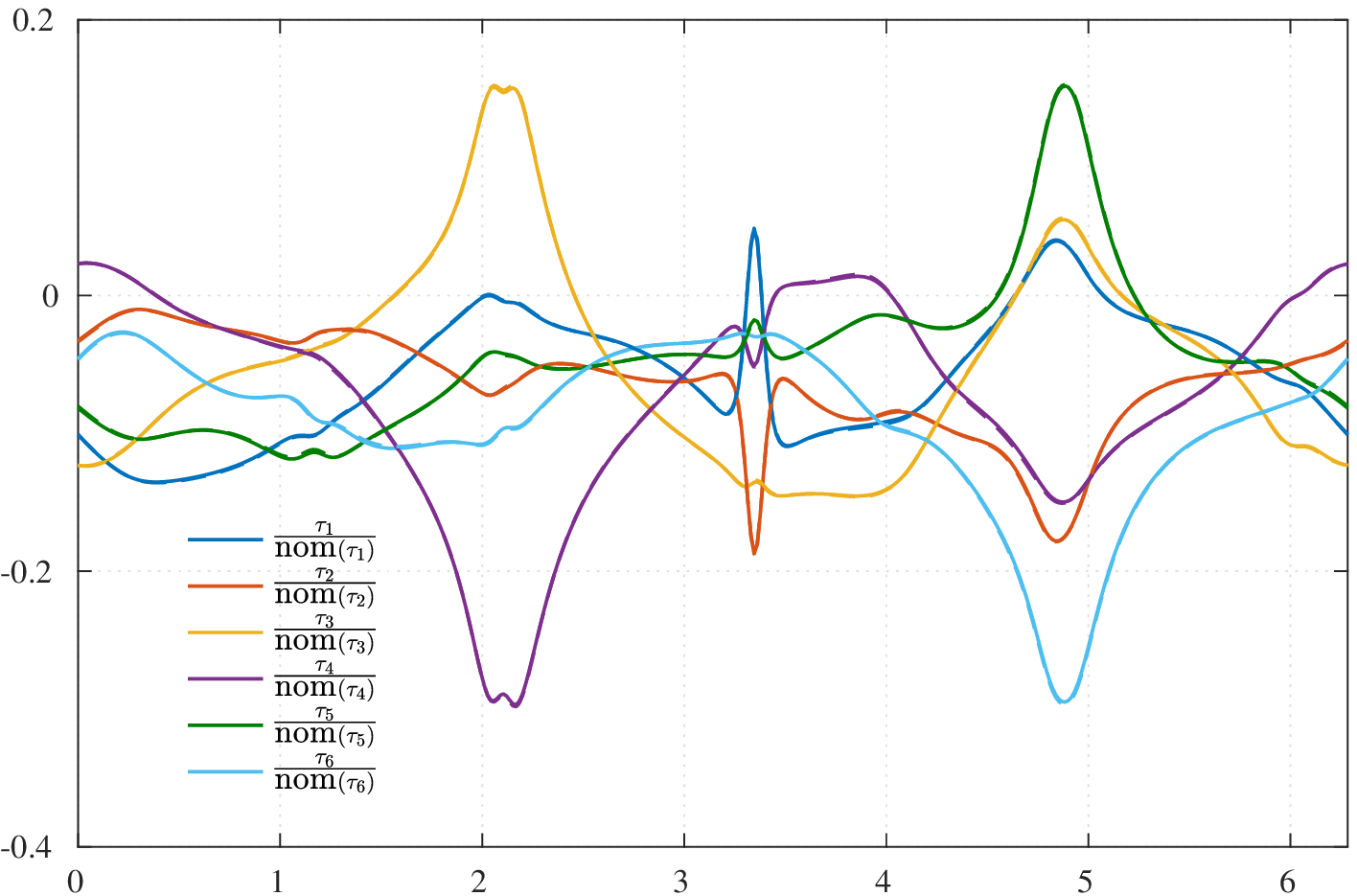} 
    \vspace*{-3mm}
    \caption{LME: Selected model (FS $n_{\vphi_{\!R}}=17$) $\frac{{\tau_i(\vphi_{\!R})}}{\mbox{nom}({\tau_i})}$ ($--$) and $\frac{{\tau_i(\vphi)}}{\mbox{nom}({\tau_i})}$ ($-$) vs. $t$ ($s$).}
    \label{fig:hexaglide_tau_norm_validation} 
  \end{center}
% \end{figure}
% %
% \begin{figure}[!ht]
  \begin{center}
    \includegraphics[width=10.32cm]{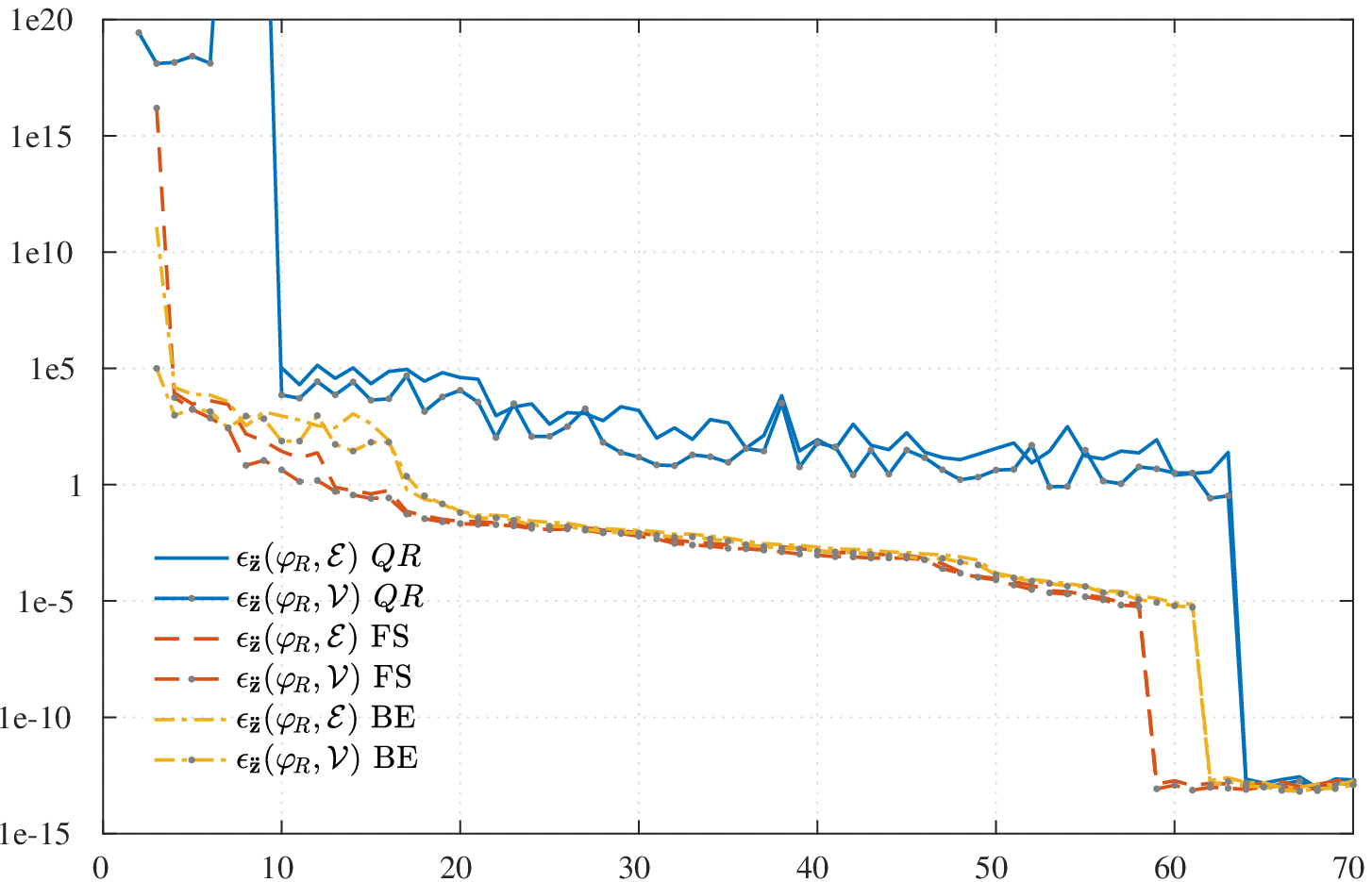} 
    \vspace*{-3mm}
    \caption{LME: $\epsilon_{\vddz}(\vphi_{\!R},{\cal E})$ and $\epsilon_{\vddz}(\vphi_{\!R},{\cal V})$ vs. $n_{\vphi_{\!R}}$ for the $QR$, FS and BE heuristics.} 
    \label{fig:hexaglide_epsilon_ddq_n_phi} 
  \end{center}
\end{figure}

For this example, the performance of the proposed algorithms is illustrated in
Figs.~\ref{fig:hexaglide_epsilon_tau_n_phi}, \ref{fig:hexaglide_n_op_n_phi}, \ref{fig:hexaglide_epsilon_tau_n_op}, \ref{fig:hexaglide_tau_norm_validation} and \ref{fig:hexaglide_epsilon_ddq_n_phi}.

Fig.~\ref{fig:hexaglide_epsilon_tau_n_phi} shows the normalized prediction error $\epsilon_{\vtau_{\!\!\vz}}$, in terms
of the number of parameters $n_{\vphi_{\!R}}$ of the parameter-reduced models, for the estimation, $\cal E$, and validation, $\cal V$, data sets based on the proposed $QR$ , FS, and BE  heuristics.
Fig.~\ref{fig:hexaglide_n_op_n_phi} plots the number of operations, $n_{op}$, required to compute $\mathbf{d}_{\!\vq}(\vq,\vdq,\vddq,\vphi_{\!R})$ and $[\M_{\!\vq\vq}|\vdelta_{\!\vq}](\vq,\vdq,\vphi_{\!R})$,
in terms of  $n_{\vphi_{\!R}}$.
Fig.~\ref{fig:hexaglide_epsilon_tau_n_op} shows the normalized prediction error $\epsilon_{\vtau_{\!\!\vz}}$ in terms of $n_{op}$ for the same functions and heuristics.
Fig.~\ref{fig:hexaglide_tau_norm_validation} compares the normalized torque for the full model and the parameter-reduced model with $n_{\vphi_R}=17$ obtained with
the FS heuristic for the validation trajectory.


Finally, for completeness, the parameter ordering corresponding to the different heuristics analyzed is presented in Table~\ref{tab::hexaglide_parameter_ordering}.
\renewcommand{\arraystretch}{1.7}
\begin{table}[!ht]
\caption{LME: Parameter ordering for the $QR$, FS and BE heuristics.}
\label{tab::hexaglide_parameter_ordering}
% \begin{minipage}[t]{0.5\linewidth}
\vspace{0pt}
\centering
\footnotesize
\begin{tabular}{ccccccccccc}
\hline
$Par. \#$ & 1 & 2 & 3 & 4 & 5 & 6 & 7 & 8 & 9 & 10  \\
\hline
BE & $m^P$ & $m^4$ & $d_z^1$ & $m^5$ & $d_z^6$ & $m^2$ & $I_{zz}^P$ & $d_z^3$ & $d_y^P$ & $I_{yy}^1$ \\
% \hline
FS & $m^P$ & $I_{zz}^P$ & $m^4$ & $d_z^1$ & $m^5$ & $d_z^2$ & $d_z^6$ & $m^3$ & $d_z^P$ & $d_x^1$ \\
% \hline
QR & $I_{xz}^P$ & $I_{xy}^P$ & $I_{yz}^P$ & $I_{xx}^P$ & $I_{zz}^P$ & $d_y^P$ & $d_x^P$ & $d_z^P$ & $I_{yy}^P$ & $d_y^4$ \\
\hline
\vspace{-0.25cm}\\
\hline
$Par. \#$ & 11 & 12 & 13 & 14 & 15 & 16 & 17 & 18 & 19 & 20  \\
\hline
BE & $I_{xx}^1$ & $m^1$ & $d_z^P$ & $I_{xx}^6$ & $I_{yy}^P$ & $I_{yy}^6$ & $I_{xx}^P$ & $I_{xy}^P$ & $I_{xx}^3$ & $I_{yy}^3$ \\
% \hline
FS & $m^2$ & $d_z^4$ & $m^6$ & $I_{xx}^1$ & $I_{yy}^P$ & $I_{xx}^P$ & $m^1$ & $I_{yy}^1$ & $d_x^6$ & $I_{xy}^P$ \\
% \hline
QR & $d_y^2$ & $d_y^6$ & $d_y^3$ & $d_y^1$ & $d_y^5$ & $d_x^1$ & $d_x^4$ & $d_z^1$ & $d_z^6$ & $d_x^5$ \\
\hline
\vspace{-0.25cm}\\
\hline
$Par. \#$ & 21 & 22 & 23 & 24 & 25 & 26 & 27 & 28 & 29 & 30  \\
\hline
BE & $d_z^5$ & $d_y^1$ & $d_y^3$ & $d_y^5$ & $d_z^4$ & $I_{yy}^2$ & $I_{xx}^2$ & $d_x^4$ & $d_x^5$ & $d_y^4$ \\
% \hline
FS & $d_x^3$ & $d_x^5$ & $d_y^1$ & $d_y^4$ & $d_y^5$ & $d_z^5$ & $d_z^3$ & $d_y^3$ & $d_x^4$ & $d_y^6$ \\
% \hline
QR & $d_x^2$ & $m^3$ & $d_z^2$ & $d_z^4$ & $d_x^3$ & $d_x^6$ & $m^5$ & $I_{xz}^4$ & $I_{xz}^1$ & $d_z^5$ \\
\hline
\vspace{-0.25cm}\\
\hline
$Par. \#$ & 31 & 32 & 33 & 34 & 35 & 36 & 37 & 38 & 39 & 40  \\
\hline
BE & $d_x^1$ & $d_y^2$ & $d_x^3$ & $d_y^6$ & $d_x^6$ & $d_x^2$ & $I_{xz}^4$ & $I_{xz}^2$ & $I_{xz}^P$ & $I_{xz}^6$ \\
% \hline
FS & $d_y^2$ & $d_x^2$ & $I_{xz}^4$ & $I_{yz}^5$ & $I_{xz}^6$ & $I_{xz}^P$ & $I_{xz}^1$ & $I_{yz}^1$ & $I_{yz}^4$ & $I_{yz}^P$ \\
% \hline
QR & $I_{xy}^1$ & $I_{yz}^4$ & $d_z^3$ & $I_{xz}^5$ & $I_{xz}^2$ & $I_{xz}^3$ & $I_{xz}^6$ & $I_{zz}^4$ & $I_{xy}^5$ & $m^1$ \\
\hline
% \end{tabular}
% \end{minipage}
% %
% \begin{minipage}[t]{0.5\linewidth}
% \vspace{0pt}
% \footnotesize
% \begin{tabular}{|c|cccccccccc|}
\vspace{-0.25cm}\\
\hline
$Par. \#$ & 41 & 42 & 43 & 44 & 45 & 46 & 47 & 48 & 49 & 50  \\
\hline
BE & $I_{yz}^P$ & $I_{xz}^1$ & $I_{yz}^1$ & $I_{yz}^4$ & $I_{xz}^3$ & $I_{yz}^6$ & $I_{xz}^5$ & $I_{yz}^2$ & $I_{yz}^5$ & $I_{yz}^3$ \\
%\hline
FS & $I_{xz}^3$ & $I_{zz}^2$ & $d_y^P$ & $I_{yz}^3$ & $I_{xz}^5$ & $I_{yz}^2$ & $I_{yz}^6$ & $I_{xz}^2$ & $I_{zz}^4$ & $I_{zz}^1$ \\
%\hline
QR & $I_{yz}^1$ & $I_{xy}^3$ & $I_{xy}^4$ & $I_{yz}^6$ & $I_{yz}^2$ & $I_{zz}^1$ & $I_{yy}^2$ & $I_{yy}^4$ & $I_{yz}^3$ & $I_{yz}^5$ \\
\hline
\vspace{-0.25cm}\\
\hline
$Par. \#$ & 51 & 52 & 53 & 54 & 55 & 56 & 57 & 58 & 59 & 60  \\
\hline
BE & $I_{zz}^4$ & $I_{zz}^1$ & $I_{zz}^5$ & $I_{zz}^3$ & $I_{zz}^2$ & $I_{zz}^6$ & $I_{xy}^1$ & $I_{xy}^4$ & $I_{xy}^5$ & $I_{xy}^2$ \\
%\hline
FS & $I_{zz}^6$ & $I_{zz}^3$ & $I_{zz}^5$ & $I_{xy}^5$ & $I_{xy}^2$ & $I_{xy}^4$ & $I_{xy}^1$ & $I_{xy}^3$ & $I_{xy}^6$ & $d_x^P$ \\
%\hline
QR & $I_{yy}^6$ & $I_{xy}^2$ & $I_{yy}^1$ & $I_{xy}^6$ & $I_{zz}^3$ & $I_{zz}^2$ & $I_{yy}^5$ & $I_{zz}^6$ & $I_{xx}^4$ & $I_{yy}^3$ \\
\hline
\vspace{-0.25cm}\\
\hline
$Par. \#$ & 61 & 62 & 63 & 64 & 65 & 66 & 67 & 68 & 69 & 70  \\
\hline
BE & $I_{xy}^3$ & $I_{xy}^6$ & $I_{xx}^5$ & $m^6$ & $m^3$ & $I_{yy}^5$ & $I_{yy}^4$ & $I_{xx}^4$ & $d_z^2$ & $d_x^P$ \\
%\hline
FS & $I_{xx}^3$ & $I_{yy}^4$ & $I_{xx}^4$ & $I_{yy}^2$ & $I_{yy}^6$ & $I_{yy}^3$ & $I_{xx}^5$ & $I_{xx}^6$ & $I_{xx}^2$ & $I_{yy}^5$ \\
%\hline
QR & $I_{zz}^5$ & $I_{xx}^2$ & $I_{xx}^6$ & $I_{xx}^5$ & $m^4$ & $I_{xx}^3$ & $I_{xx}^1$ & $m^2$ & $m^6$ & $m^P$ \\
\hline
\end{tabular}
% \end{minipage}

\end{table}
 \renewcommand{\arraystretch}{1}
 
\afterpage{\clearpage}

\pagebreak
%
%--------------------------------------------------
\section{Discussion} \label{sec::discussion}

In this section, the results of the previously  presented simulations are discussed. An interpretation of the results is given, which aims to draw conclusions that can
be extended to a more general MSD context.
First, we analyze the reduced parameter models in terms of their accuracy in the range in which the
error is negligible. Then, the same analysis is performed for larger reductions with an appreciable effect on precision. Finally, the  computational  time savings for the
different parameter-reduced models are analyzed.

\subsubsection*{Parameter reduction without loss of accuracy}

For a sufficiently exciting estimation trajectory, $\cal E$,
the rank of the observation matrix, $r=\mbox{rank}(\W({\cal E}))$, is maximized. In this case, this rank coincides
with the number of base parameters, which is $36$ and $64$ for the analyzed HME and LME, respectively.
%For a sufficiently exciting estimation trajectory $r=\mbox{rank}(\W)$ is maximized and is coincident with the number of base parameters:  $36$  and $64$ respectively.

For both examples, there are ridges in the plots for $\epsilon_{\vtau_{\!\!\vz}}$ vs. $n_{\vphi_{\!R}}$ for all the heuristics: $QR$, FS and BE.
These ridges are clearly appreciable in Figs.~\ref{fig:puma560_epsilon_tau_n_phi} (HME) and \ref{fig:hexaglide_epsilon_tau_n_phi} (LME).
For the $QR$ heuristic, as expected, the right of this ridge
coincides exactly with the number of base parameters. This confirms that the estimation trajectories used in this work are sufficiently exciting and, therefore, appropriate to support
the analyses.
The models to the right of these ridges  will be said to have ``full precision'', as
the small noise appreciable in this zone is related to numerical round-off errors and not to the reduced parameter number.

 In the same figures, both for  the HME and the LME, it is important to notice that some of the ridges for the FS and BE heuristics are displaced to the left
relative to those of the $QR$ heuristic. This can be explained by the fact that the FS and BE heuristics optimize the prediction error $\epsilon_{\vtau_{\!\!\vz}}$, while $QR$ minimizes
the estimation error for the model parameters. For the HME
only the FS heuristic achieves full precision for a model with a number of parameters, $n_{\vphi_{\!R}}$, that is smaller than $r$, $n_{\vphi_{\!R}}=r-1=35$. In the case of the LME, a
reduction in the minimum number of parameters required to achieve full precision
is appreciated for both the FS and BE heuristics.
The FS heuristic clearly performs best followed  by BE, requiring a minimum of $n_{\vphi_{\!R}}=r-5=59$ and $n_{\vphi_{\!R}}=r-2=62$ parameters, respectively.
The fact that the distance of the FS and LS ridges to the left of the  $QR$ ridge is larger in the LME than in the HME is a direct consequence of its ``low mobility''.
%\footnotemark.
%\footnotetext{To see this in perspective note that a $\pi/2$ rotation of the axis of the first rotation on the U joint removes $6$ base parameters.}
% In addition to this, for mechanisms with closed loops as the Hexaglide, the dynamic equations are initially written
% based on a loop tree structure that has associated a bigger number of base parameters ($70$ in our case). That is why  for systems with closed loops the reduction to base parameters
% is expected to have a bigger impact on the operation count.

These ridges  also appear in the plots for  $\epsilon_{\vddz}$  vs. $n_{\vphi_{\!R}}$ shown in Figs.~\ref{fig:puma560_epsilon_ddq_n_phi} (HME) and \ref{fig:hexaglide_epsilon_ddq_n_phi} (LME).
For all the heuristics, $QR$, FS and BE, these ridges  are placed exactly at the same number of parameters, $n_{\vphi_{\!R}}$,
as in the $\epsilon_{\vtau_{\!\!\vz}}$ vs. $n_{\vphi_{\!R}}$ plots discussed in the previous paragraphs. So the same conclusions
about the attainable reduction in the number of parameters, while maintaining full precision, are applicable not only to
 the IDM but also to the DDM.
 
In the light of the aforementioned similarities, noting that the results in this paper are based on the minimization of
 $\epsilon_{\vtau_{\!\!\vz}}$, it seems that a hypothetical
 model selection based on the minimization of $\epsilon_{\vddz}$,  instead of the minimization of $\epsilon_{\vtau_{\!\!\vz}}$,  will produce similar results for parameter-reduced models with full precision.

\subsubsection*{Parameter reduction with small loss of accuracy}

Now, attention is focused on the parameter-reduced models on the left of the ridges described above. In this zone, a loss of accuracy which
 is not occluded by numerical round-off errors  can be observed.

Going back to the plots for $\epsilon_{\vtau_{\!\!\vz}}$ vs. $n_{\vphi_{\!R}}$ shown in Figs.~\ref{fig:puma560_epsilon_tau_n_phi} (HME) and \ref{fig:hexaglide_epsilon_tau_n_phi} (LME),
it can be seen that the proposed FS and BE heuristics outperform
 the $QR$ heuristic, showing errors of a  smaller order of magnitude for the same number of reduced parameters. In fact, the $QR$ heuristic cannot be claimed to be a
 reasonable model selection or parameter reduction method on the left of its ridge. As before, this is explained because
 the $QR$ heuristic minimizes the error on the estimation of the parameters and not the prediction error $\epsilon_{\vtau_{\!\!\vz}}$.
 This puts the relevance of the performance indexes chosen to obtain parameter-reduced models into perspective.
 
These figures clearly show that the FS and BE heuristics can be used to obtain \textit{very significant}
reductions in the number of parameters while still keeping a respectable level of accuracy. Based on the FS heuristic,
errors of about $\epsilon_{\vtau_{\!\!\vz}}\approx 10^{-2}$ can be obtained for the HME and the LME with as few as  $n_{\vphi_{\!R}}=18$ and $n_{\vphi_{\!R}}=17$  parameters, respectively.
These are the parameter-reduced models selected for discussion throughout this section. See Table~\ref{multiprogram}, which summarizes the relevant data used in the discussion of these examples.

Even larger reductions in the number of parameters can be obtained if precision requirements are relaxed a bit more. For a given  $n_{\vphi_{\!R}}$, at the left of the corresponding ridge, the FS and FE heuristics
give a similar $\epsilon_{\vphi_{\!R}}$ error. Nevertheless, FS  consistently performs better
than  BE in both examples.

In the same zone, the plots for $\epsilon_{\vddz}$  vs. $n_{\vphi_{\!R}}$ shown in Figs.~\ref{fig:puma560_epsilon_ddq_n_phi} (HME) and \ref{fig:hexaglide_epsilon_ddq_n_phi} (LME) depict the same tendency:
parameter-reduced models with
the same number of parameters as before, $n_{\vphi_{\!R}}=18$  for the HME and $17$ for the LME, show acceptable $\epsilon_{\vddz}$ error values (see Table~\ref{multiprogram}). This
error seems to increase rapidly if the number of parameters
is further reduced. 
In both examples, even for the best-performing FS heuristic,  errors $>100\%$ for a parameter count of $\approx 10$ can be observed. 
This sharp increase is related to the degradation of the condition number of the mass matrix $\M_{\!\vz\vz}$, whose inverse is required to compute $\vddz$.
Note that this error goes up to $+\infty$ when $\M_{\!\vz\vz}$ becomes singular.
This suggests that in order to improve the performance of the parameter-reduced DDM,
 $\epsilon_{{\vtau}_{\vz}}$ can be substituted as the optimization criterion by $\epsilon_{\vddz}$. This may be of particular interest when dealing with parameter-reduced models with a small number of parameters.
An interesting benefit of using this error as the optimization criterion is that it does not become singular for underactuated multibody systems.

Again it is seen that, for an equal number of parameters, the FS heuristic shows smaller errors than the BE heuristic. Let us use the LME to explain this difference:
the FS heuristic algorithm begins with the full set of parameters, $\vphi_{\!R}=\vphi$. When it starts removing parameters, it faces the removal of parameters
that have a negligible effect on the dynamics.
With $n_{\vphi_{\!R}}$ between $70$ and $r=\mbox{rank}(\W({\cal E}))=64$, at every reduction step
the removal of parameters has no effect at all on the dynamics. There is a large number of parameters that can be chosen to be removed, and the algorithm randomly decides
which one because, in this range, $\epsilon_{\vtau_{\!\!\vz}}$ is just numerical noise. But this initial choice of removed parameters affects
 the parameter-reduced models obtained afterward. 
On the other hand, the FS algorithm starts with an empty set of parameters. At each step, it always adds
parameters with a non-negligible contribution to the dynamics and, even if it is not optimal, it will always capture the leftmost ridge position.
As the results show, this makes a difference. 
A similar reasoning can be applied to the HME.
These results are in accordance with results observed for the FS and BE heuristics in other scientific contexts \cite{Nelles2001}.

\subsubsection*{Operation count savings vs precision}

 The number of operations in relation to the number of parameters, $n_{op}$ vs. $n_{\vphi_{\!R}}$, for the different heuristics is shown in Figs.~\ref{fig:puma560_n_op_n_phi} (HME) and \ref{fig:hexaglide_n_op_n_phi} (LME).
It is important to note that the criteria used for
model selection do not include minimization of the operation count of the reduced models.
Nevertheless, an important decay in the number of operations is observed as the number of parameters in the model is reduced. This decay exhibits an almost piece-wise linear profile, with
  slopes that increase with the number of parameters removed.
The figures suggest that there are some advantages to the FS and BE heuristics in comparison with the $QR$ heuristic. It is important to remember that
the operation count is dependent on different aspects, such as the parametrization used or the formalism and the way in which it is implemented.
As mentioned in Section~\ref{sec::results}, in  both examples state-of-the-art general purpose symbolic formulations \cite{Samin2003,Ros2007} have been used to obtain the model functions
and, therefore, the operation count. In this way, the results presented give a correct perspective on the real savings attainable using the proposed methods.

The normalized prediction error as a function of the number of operations, $\epsilon_{\vtau_{\vz}}$ vs. $n_{op}$,  for the different heuristics is presented in Figs.~\ref{fig:puma560_epsilon_tau_n_op} (HME) and \ref{fig:hexaglide_epsilon_tau_n_op} (LME).
Note that the information given in these plots is just a convenient rearrangement of the information  given in Figs.~\ref{fig:puma560_epsilon_tau_n_phi} and \ref{fig:puma560_n_op_n_phi} (HME)
and in Figs.~\ref{fig:hexaglide_epsilon_tau_n_phi} and \ref{fig:hexaglide_n_op_n_phi} (LME), respectively.
This is a more useful representation of the data if a balance between the precision and computational requirements of the chosen parameter-reduced model is sought.
From these figures it can be observed that, for the previously selected models (FS with $n_{\vphi_{\!R}}=18$ for HME and  FS with $n_{\vphi_{\!R}}=17$ for the LME,
with a prediction error of $\epsilon_{\vtau_{\!\!z}}\approx 10^{-2}$), the reductions in the computational cost for computation of the  IDM and DDM functions are above $46\%$ in the case of the HME and above
$67\%$ in the case of the LME (see Table~\ref{multiprogram}).

\begin{table}[H]
        \caption{Results for selected  HME \& LME  models: heuristic H, parameter number $n_{\vphi}$, base parameter number $r$, reduced parameter number $n_{\vphi_{\!R}}$,
        and full and reduced op. number $n_{op}(\vphi)$ and $n_{op}(\vphi_{\!R})$, $n_{op}(\vphi_{\!R})/n_{op}(\vphi)$ ``red.'' and normalized errors $\epsilon_{\vtau_{\!\!\vz}}(\vphi_{\!R})$ and $\epsilon_{\vddz}(\vphi_{\!R})$.
        Number of operations and error results are given for the IDM ($\mathbf{d}$) and DDM ($[\M,\vdelta]$).}
        \centering
        \label{multiprogram}
        \begin{tabular}{ccccccccccccccc}
%         \begin{tabular}{ccc|cccc|cccc|}
            %\cline{4-11}
             && &&&& \multicolumn{4}{c}{$\mathbf{d}$} & ~&\multicolumn{4}{c}{$[\M,\vdelta]$} \\             %\cline{2-9}
\cline{2-5} \cline{7-10}\cline{12-15}
         & H &    $n_{\vphi}$  & $r$ &  $n_{\vphi_{\!R}}$ & ~ & $n_{op}(\vphi)$ & {$n_{op}(\vphi_{\!R})$} &  red. & $\epsilon_{\vtau_{\!\!\vz}}(\vphi_{\!R})$ & ~ & $n_{op}(\vphi)$ & $n_{op}(\vphi_{\!R})$ &  red. &  $\epsilon_{\vddz}(\vphi_{\!R})$  \\
         %\hline
            HME & FS  &   49 & 36 & $18$ & ~& 719 & 460  &  36\%  &  1.3\% & ~  & 1658 & 1039  &  37\%  &    4.7\%\\ 
            LME & FS  &   70 & 64 & $17$ & ~& 1030 & 331  &  68\%  &  0.8\% & ~  &  1007 & 330  &  67\%  &    5.4\%\\ 
%              \hline
\cline{2-5} \cline{7-10}\cline{12-15}
        \end{tabular}
    \end{table}

In a more direct way, Figs.~\ref{fig:puma560_tau_norm_validation}  (HME) and \ref{fig:hexaglide_tau_norm_validation} (LME) compare the normalized values
of the actuator moments and forces along the validation trajectory for the
full and selected parameter-reduced models  (FS with $n_{\vphi_{\!R}}=18$ for HME and FS with $n_{\vphi_{\!R}}=17$ for LME), respectively.
As can be seen, the trajectories are barely distinguishable by the naked eye.

% For completeness, the parameter ordering corresponding to the different heuristics analyzed is presented in Tables~\ref{tab::puma560_parameter_ordering}
% and \ref{tab::hexaglide_parameter_ordering}, respectively. It has been observed that the number of points considered on the
% estimation trajectories has some influence in the parameter ordering obtained, results are not shown due the lack of space. As very important parameter
% reductions have been shown, this sensibility indicates that there is still room for improvement. In particular, the introduction of optimization
% strategies that explicitly penalize parameter-reduced model operation count are expected to improve the results presented. But also, more involved
% parameter selection procedures that check for a bigger number of parameter combinations, as the
% one suggested in section \ref{sec::other_forward_backward_heuristics}, deserve further investigation.

\section{Conclusions} \label{sec::conclusions}

In this paper, drawing inspiration from the literature on model selection, parameter reduction methodologies have been proposed and tested in the context of MSD models.
These methods require: a) a suitable selection heuristic, b) a characteristic data set representative of the dynamics of the system, and c) a performance measure 
for the parameter-reduced models.

Three different model selection heuristics -$QR$ decomposition-based ($QR$), \textit{Forward Selection} (FS) and \textit{Backward Elimination} (BE)- have been proposed and tested in two
multibody systems with different -high (HME) and low(LME)- characteristic mobility. A normalized error for the inverse dynamics model
is defined and used as the performance measure in these examples. A detailed description of the systems, procedure and results is given.

The results show that, based on the FS heuristics and the proposed performance measure,
an impressive reduction in the number of parameters required is obtained:
for a normalized error for the inverse dynamics model of $\approx1\%$  only $18$ parameters out of $49$ for the HME  and
$17$ out of $70$ for the LME, are required. Corresponding to these parameter-reduced models, computational complexity savings for the evaluation of the functions required
for the direct and inverse dynamics models are above $36\%$ for the HME  and above $67\%$ for the LME. 
A defined normalized error for the direct dynamics model shows errors of $\approx5\%$ for these same models.

A detailed description and interpretation of the results is given. In general, the FS heuristic, closely followed by BE, appears
to be the best-performing one for any number of parameters. $QR$, a good heuristic to reduce the error in the parameter estimation context,
 is discarded due to inadmissible levels for the proposed normalized errors.
 The degradation of the direct dynamics seen for small parameter-reduced models is related to the
worsening of the condition number of the mass matrix.

For any reduction level, the better performance of the methods for the LME is seen to be related to its low mobility.
Applications of MSD for low-mobility systems  are likely to benefit especially from the methods presented. Important cases in this context
are railway and automotive dynamics. For these, the methods proposed can offer very large reductions in the number of parameters, as they present low mobility.
This leads to more robust parameter estimation procedures and much smaller computational costs of the IDM and DDM, offering new possibilities to
 applications of MSD that require real-time performance.
 
As an additional conclusion, it has been shown that it is possible to obtain analytical expressions that determine the value of the estimated reduced set of parameters
as a function of the parameter values of the original model. These analytical expressions can be considered a generalization of the  \emph{base parameter} concept
to the case of approximate parameter-reduced models.

\subsubsection*{Future work}

Two lines have been identified as having research potential: first, using the  normalized error for the direct
dynamics as the optimization criterion for further improving the quality of direct dynamics parameter-reduced models with a small
number of parameters. Second,  within computational resources, widening the search space of FS-style heuristics to obtain even better results. 

With regard to the second line, it is interesting to mention the important case of multibody systems models
exhibiting symmetries. The space search can be reduced enormously if characteristic data sets exhibiting the same symmetry are used and if
dynamic parameters with a symmetric dynamic role are selected or eliminated in groups. This will generally make it possible to obtain better
results, or even optimal results in some cases.

\section*{Acknowledgements}
 
This work was partially supported by the ``Plan Nacional de I+D, Comisi\'on Interministerial de Ciencia y Tecnolog\'ia (FEDER-CICYT)'' [grant number DPI2013-44227-R];
and by the Spanish Ministry of Economy and Competitiveness (MINECO) [grant number TRA2014-57609-R].




  \bibliographystyle{elsarticle-num}

%  \section*{\refname}

  \bibliography{bibliografia}





% \end{thebibliography}
\end{document}